\documentclass[twoside,11pt]{article}
\usepackage[table, dvipsnames]{xcolor}
\usepackage{bm}
\usepackage{float}
\usepackage{cite}
\usepackage{appendix}
\usepackage[margin=1in]{geometry}
\usepackage{amsfonts, amssymb, amsmath, amsthm, multicol}
\usepackage[colorlinks=true, pdfstartview=FitV, linkcolor=blue,
citecolor=blue, urlcolor=blue]{hyperref}
\definecolor{Red}{rgb}{1,0.05,0}
\definecolor{Grn}{rgb}{0.1,0.7,0.1}
\definecolor{Blu}{rgb}{0.1,0.1,0.6}
\definecolor{Org}{rgb}{1,0.45,0}
\definecolor{Vio}{rgb}{0.6578,0,0.9478}
\definecolor{Mag}{rgb}{1,0.2,0.3}
\usepackage{authblk}
\usepackage{breqn}
\usepackage{mathtools}
\DeclareMathOperator{\acosh}{acosh}
\usepackage{cleveref}
\usepackage{multirow}
\usepackage{gensymb}
\usepackage{pdflscape}
\usepackage{footnote}
\usepackage{lscape}
\makesavenoteenv{tabular}
\newcolumntype{C}[1]{>{\centering\arraybackslash}p{#1}}
\newcolumntype{L}[1]{>{\raggedright\arraybackslash}p{#1}}
\newcolumntype{M}[1]{>{\centering\arraybackslash}m{#1}}
\usepackage[mathlines]{lineno}
\hypersetup{
    colorlinks,
    linkcolor={red!50!black},
    citecolor={blue!50!black},
    urlcolor={blue!80!black}
}
\usepackage{graphics}
\usepackage{tikz}
\usepackage{hyperref}
\usepackage{overpic}
\usepackage{caption}[=v3.6]
\usepackage{subcaption}
\usepackage{amssymb}
\usepackage{amsmath}
\usepackage{array}
\usepackage{rotating}
\usepackage{graphicx}
\usepackage[export]{adjustbox}
\usepackage{makecell}

\hypersetup{
    colorlinks,
    linkcolor={red!50!black},
    citecolor={blue!50!black},
    urlcolor={blue!80!black}
}

\usepackage{soul}

\makeatother

\graphicspath{{images/}}
\newcommand{\rotcentered}[1]{\adjustbox{rotate=90,valign=m}{#1}}

\NewDocumentCommand{\suboverpic}{O{.9} O{black} O{1,50} m O{#4}}{
    \adjustbox{valign=M}{
        \begin{subfigure}{#1\linewidth}
            \centering
            \phantomcaption
            \label{fig:#5}
            \begin{overpic}[width=\linewidth]{images/#4.png} 
            \put(#3){\color{#2}{(\thesubfigure)}}
            \end{overpic}
        \end{subfigure} 
            }
}

\numberwithin{rmk}{section}
\numberwithin{nt}{section}

\title{Breaking the Ice: Video Segmentation of Close-Range Ice-Covered Waters}

\author[1, a]{Corwin MacMillan}
\author[1, a]{K. Andrea Scott}
\author[2, b]{Matthew Garvin}
\author[1, a]{Zhao Pan}
\affil[1]{University of Waterloo, Department of Mechanical and Mechatronics Engineering, Waterloo, ON, Canada}
\affil[a]{{\{cgjmacmi, ka3scott, zhao.pan\}@uwaterloo.ca}}
\affil[2]{Canada National Research Council, St John's, NL, Canada}
\affil[b]{matthew.garvin@nrc-cnrc.gc.ca}

\date{\today}

\begin{document}
\maketitle \vspace{1cm}

\section*{Abstract}
Rapid ice recession in the Arctic Ocean, with predictions of ice-free summers by 2060, opens new maritime routes but requires reliable navigation solutions. 
        Current approaches rely heavily on subjective expert judgment, underscoring the need for automated, data-driven solutions.
        This study leverages machine learning to assess ice conditions using ship-borne optical data, introducing a finely annotated dataset of 946 images, and a semi-manual, region-based annotation technique.
        The proposed video segmentation model, UPerFlow, advances the SegFlow architecture by incorporating a six-channel ResNet encoder, separate UPerNet-based segmentation decoders for each image, PWCNet as the optical flow encoder, and cross-connections that integrate bi-directional flow features without loss of latent information.
        The proposed architecture outperforms baseline image segmentation networks by an average 38\% in occluded regions, demonstrating the robustness of video segmentation in addressing challenging Arctic conditions.

\clearpage

\section{Introduction} \label{intro}
\begin{figure*}[t]
\centering
    \adjustbox{max width=\linewidth, center}{
     \suboverpic[.3][black][1,49]{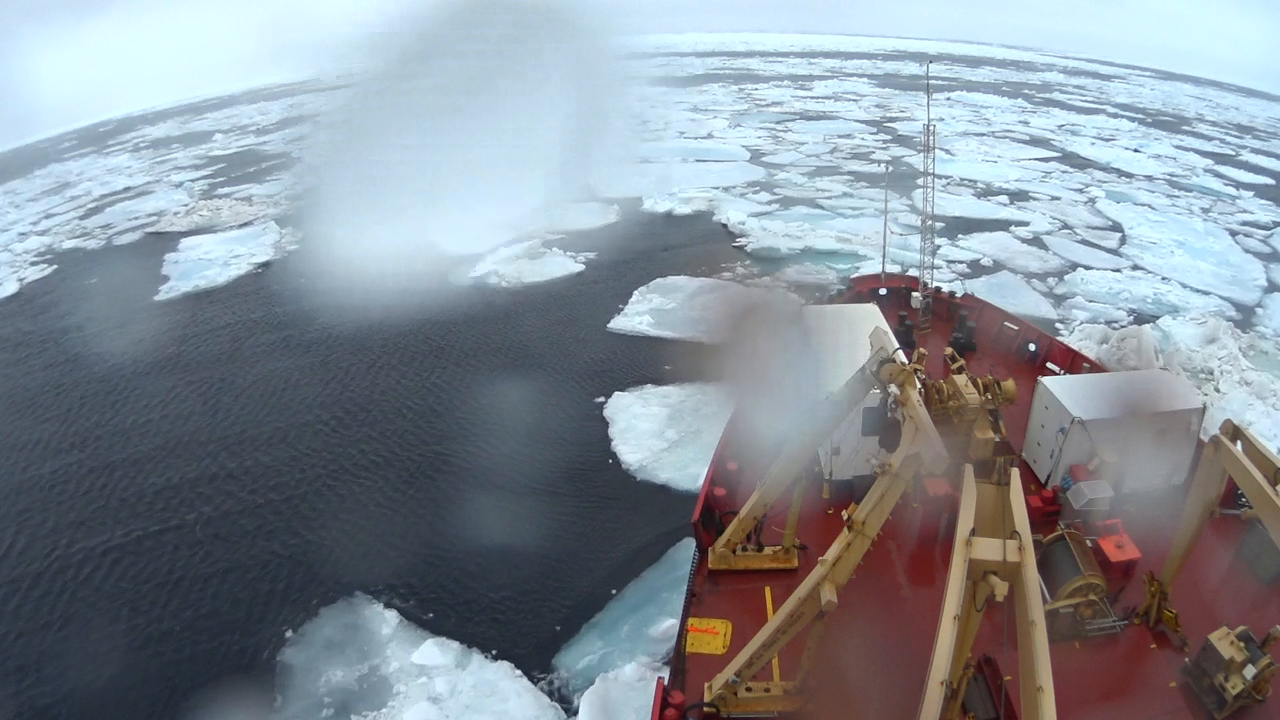}[unlabelrain] \hfill
     \suboverpic[.3][black][1,49]{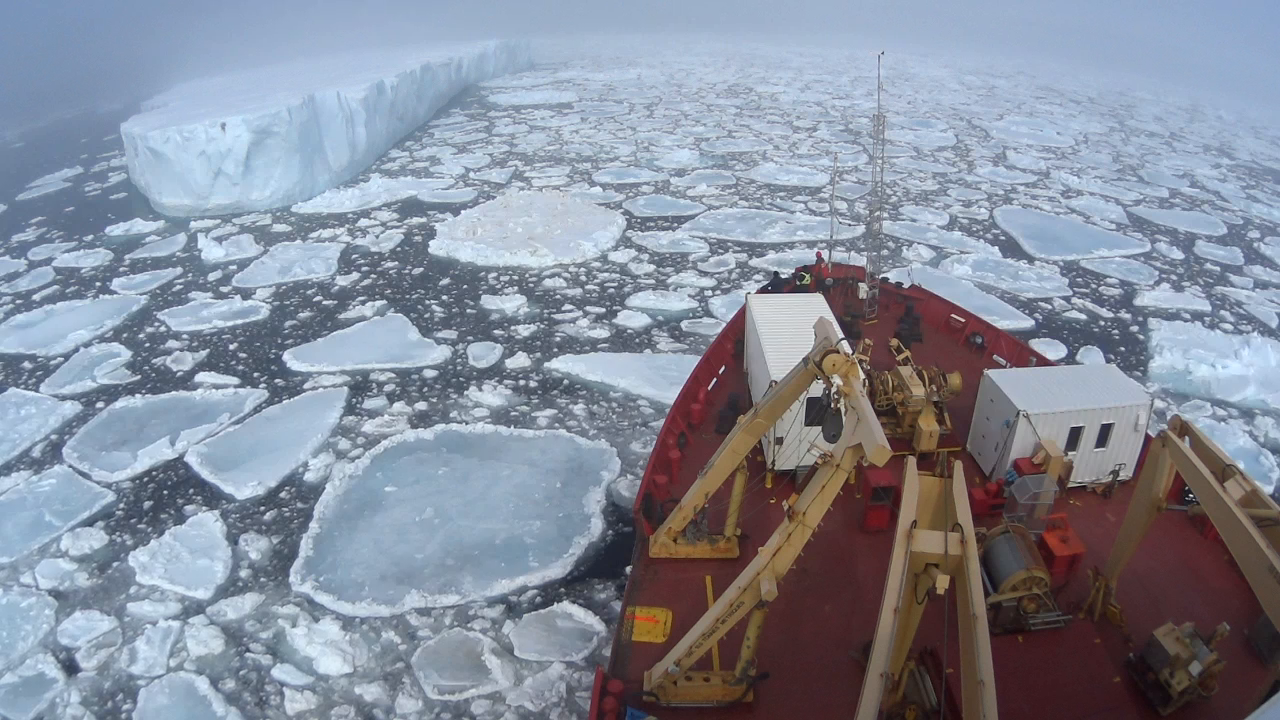}[valframe360] \hfill
     \suboverpic[.3][yellow][1,49]{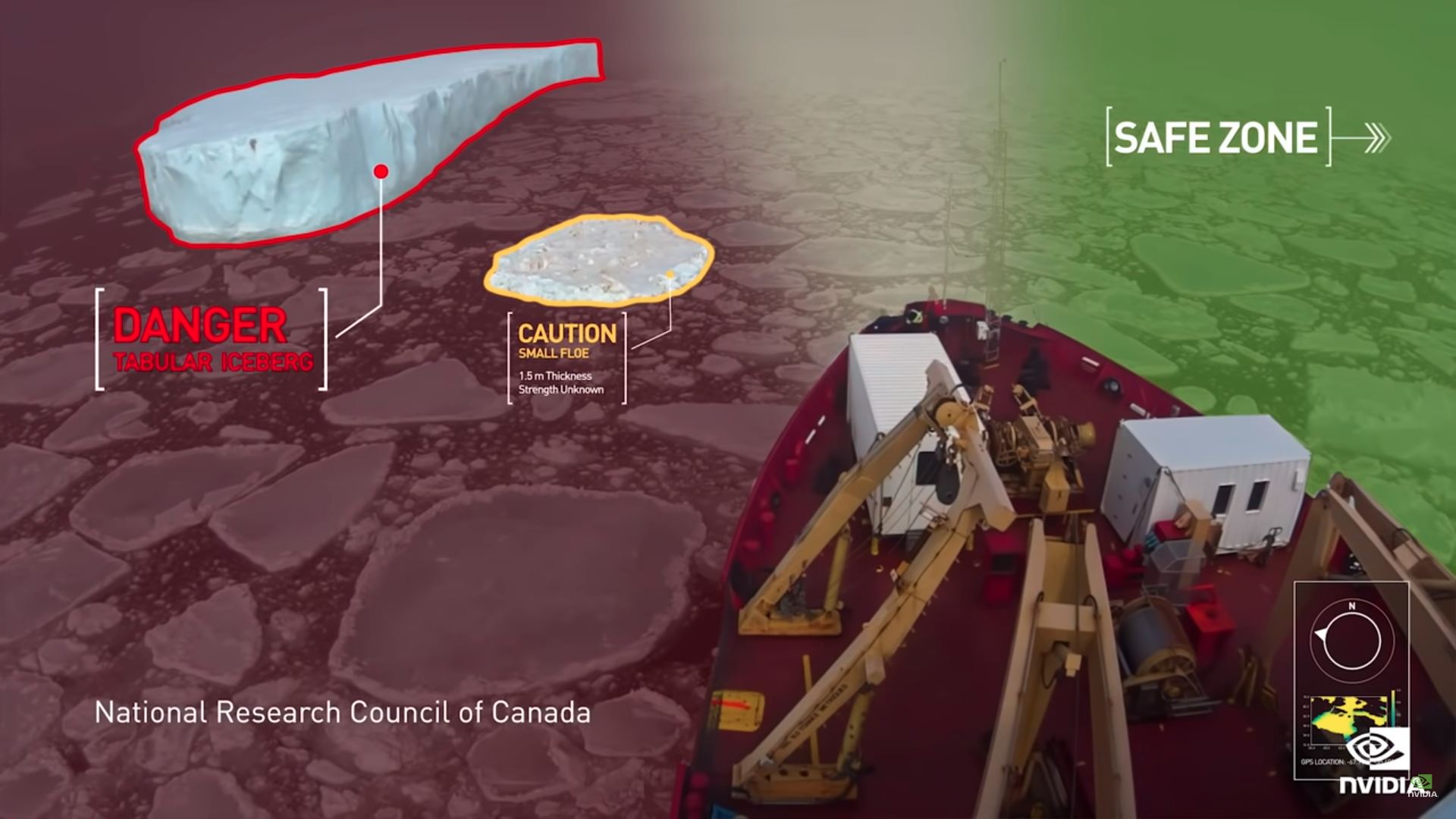}[nvidialabel]
     }
\caption{Examples of images from the Amundsen dataset. \subref{fig:unlabelrain} shows a common occlusion in the data, where a water droplet on the lens partially obscures the image. \subref{fig:valframe360} shows an example from the validation dataset. \subref{fig:nvidialabel} shows \subref{fig:valframe360} as it was shown in NVIDIA's 2021 keynote~\cite{NVIDIAkeynote}.}
\label{fig:exampledataset}
\end{figure*}

Arctic sea ice has shown a rapid and irrefutable trend of coverage recession and thinning over recent decades. It is projected that ice-free conditions in the summer can be expected by as early as 2060 \cite{2060iceconditions}. While this has negative implications, especially as they pertain to the environment, there also exist some that are positive, including improved accessibility of the Arctic for navigation. Of particular interest is the opportunity to use the Arctic as a trade route between the Pacific and Atlantic, providing alternatives to the currently relied upon Panama Canal through the Northern Sea Route, the transpolar route, and eventually the Northwest passage \cite{ostreng2013shipping}. For many, trade through the Arctic may be the most attractive and convenient option. For example, transportation between Yokohama, Japan and Rotterdam, Netherlands is reduced from 20,700 kilometers through the Panama Canal to 13,500 through the Northwest Passage, equating to a 40\% reduction in time, carbon dioxide emissions, and fuel costs. Currently, ice navigation relies on experts to interpret and integrate data from tools such as ship radar and satellite imagery with their observations to make subjective decisions about safe navigation conditions. There exists a need for tools that can assist these experts by providing objective, data-driven assessments of navigational safety.

There are several factors that determine the overall ``safeness'' of the ice conditions surrounding a ship vessel. 
Intuitively, factors such as the amount of ice coverage and the size of ice pieces have a direct correlation to the safeness of the ice condition. 
Additional factors such as the age of the ice (e.g., first-year ice versus multiyear ice), and the pressure caused by the movement and joining of ice floes pose perhaps a larger threat to the safety of the ship~\cite{garvin2020review}.
To properly assess these factors, researchers seek to use various tools to create an objective measure of the ice condition, including ice radar, accelerometers, and LiDAR (Light Detection and Ranging)~\cite{lund2018radar, heyn2020accel, petty2016lidar}. 

This work is interested in assessing the ice situation using data from shipborne cameras; an increasingly popular approach in the domain. These data can provide a large field of view at a high-resolution and high frame-rate, to provide dense spatial-temporal information on the ice conditions surrounding the ship. To classify the different ice objects in the images, unsupervised segmentation techniques, including K-means clustering, Otsu's method, and thresholding have been used \cite{remund1998kmeans, coggins2014kmeans, fuvckar2016kmeans, sobiech2013kmeans, bharathi2013kmeans, WEISSLING2009eiscam, SANDRU2020kmeans, zhang2014otsukmeans, desilva2016iceunsupervised, wenjun2016unsupervisedconcentration, kalke2018svm}. 
These unsupervised techniques operate without prior knowledge of the underlying image content and rely solely on statistical properties of the image data, which may not always correspond to meaningful distinctions between ice and non-ice regions. 
Additionally, they are sensitive to variations in image characteristics, such as illumination conditions, contrast levels, and noise levels. As a result, they often produce inconsistent and inaccurate results, particularly in complex scenes with heterogeneous ice cover and diverse surface conditions. This leads to significant reliability issues that limit their effectiveness in practical applications.  

Neural networks are able to address the lack of contextual information of unsupervised techniques and learn multi-dimensional patterns from complex data. Convolutional neural networks (CNNs), that learn convolutional weights to capture spatial information in an image \cite{li2020cnnsurvey}, are well suited to optical sea-ice imagery.
Related work on the segmentation and classification of ship-borne optical ice data use well-established techniques in the field of semantic segmentation \cite{dowden, icerainremoval, icegan, dowdenfaster, icedeeplabattention, icedeeplabattention2, iceunetmodifications, panchi, panchi2024weatherdegraded}. 
However, these works share some limitations. 
Since there are no public datasets for sea ice segmentation, researchers annotate their own datasets, which tend to be small or use coarse label annotations. 
Furthermore, there is limited work on the mitigation of erroneous predictions from lens occlusions, such as water droplets on the lens. 
Finally, there are no published studies that leverage temporal information in order to improve accuracy in occluded regions. 
Based on these observations, the contributions of this study are:

\begin{enumerate}
    \item A novel labeling methodology leading to the development of a medium-sized finely annotated dataset classifying ice floes, brash ice, water, ship, sky, and iceberg.
    \item The assessment of pre-existing image semantic segmentation networks, establishing a baseline of accurate classifiers for ice situation assessment.
    \item Improvement upon existing video semantic segmentation architectures to leverage temporal information and improve performance overall and in occluded regions.
\end{enumerate}

This paper is organized as follows: Section~\ref{background} outlines related works interested in close-range ice data annotation and segmentation. Section~\ref{dataset} outlines the dataset and labeling methodology. Section~\ref{methodology} introduces UPerFlow, an optical flow-based video segmentation model, and the training and evaluation methodology.
Finally, results and discussion are provided in Section~\ref{results}.

\section{Background} \label{background}

This section provides a comprehensive background for the key aspects relevant to the present work. A brief literature review is provided for works as they pertain to close-range sea ice datasets, unsupervised segmentation, semantic segmentation neural networks, as well as optical flow, and video segmentation.

\textbf{Close-range sea ice datasets.} 
Annotating close-range sea ice imagery is challenging due to limited publicly available datasets and the high cost of labeling images. The choice is often between resource-intensive manual annotation or faster but less reliable unsupervised methods. Dowden et al.~\cite{dowden} manually annotated video data from the Nathaniel B. Palmer Antarctic expedition~\cite{nathanielbpalmer}, resulting in a dataset of 1,090 images. Their dataset contains some coarse annotations, such as fields of ice floes being grouped as single entities. Asharay et al.~\cite{icerainremoval} focused on enhancing model performance through raindrop removal. Their dataset contained less than half the number of annotated images as Dowden et al. While numerous studies have built on the dataset of \cite{dowden} or developed their own~\cite{icegan, dowdenfaster, icedeeplabattention, icedeeplabattention2, iceunetmodifications, panchi}, issues of coarse annotations and small dataset sizes persist. The present study provides a practical alternative by combining unsupervised techniques with manual annotation, creating a mid-sized, high-resolution dataset of about 1,000 images. 

\textbf{Unsupervised segmentation for close-range sea ice data.}
Global segmentation techniques such as K-means clustering have been effectively used for close-range applications~\cite{WEISSLING2009eiscam, SANDRU2020kmeans} of sea ice imaging. 
However, global techniques are limited to ``clean'' data that exhibit consistent properties throughout the image (e.g. uniform lighting conditions).
For close-range sea ice data, global approaches can often be naive, since atmospheric scattering of light causes distribution of bright pixels to vary with distance to the camera. 
To address these challenges, this study proposes segmenting images into multiple regions based on distance, allowing for tailored segmentation in each region. This method adapts to the varying conditions across the image, overcoming the limitations of a single global model.
Otsu's algorithm~\cite{otsu}, which optimizes thresholds based on pixel intensity histograms, has been effective in both remote~\cite{zhang2014otsukmeans} and close-range applications for its simplicity and efficiency~\cite{desilva2016iceunsupervised, wenjun2016unsupervisedconcentration}.
The present work extends Otsu's method by using multiclass thresholding~\cite{arora2008multilevel} to generate several proposals for the correct segmentation, from which the best-fit threshold is chosen manually for improved accuracy. 

Similar to this annotation process is the work of Haverkamp et~al.~\cite{haverkamp1995localdynamicthresholding} and Lu et~al.~\cite{lu2010shipthresholding}, who separate ice imagery into multiple regions and apply unsupervised techniques in each region. 
In the work of Lu et~al., they focus on close-range ice concentration, and discard far-field information in favor of higher accuracy near-field segmentations. 
The present study argues that far-field information is still important and should be used for navigation and machine learning. 
Although individual ice floes may not be discernible in the far-field, pertinent information about concentration or the presence of dangerous icebergs can still be estimated.

\textbf{Neural networks for segmentation of close-range sea ice imagery }
Advancements in deep learning have led to the development of semantic segmentation networks, which learn to recognize and classify pixel-level information. These networks, such as fully convolutional networks (FCNs)~\cite{long2015fcn} and more sophisticated architectures like U-Net~\cite{ronneberger2015unet}, the DeepLab series~\cite{chen2018deeplabv3plus}, PSPNet~\cite{zhao2017pspnet}, or UPerNet~\cite{xiao2018upernet}, leverage the ability of CNNs to extract hierarchical features from images, making them well-suited for the detailed segmentation tasks required for ice-covered waters.

Dowden et al.~\cite{dowden} first showed the potential of these approaches for ice classification and detection in ship data. 
They used PSPNet to achieve 0.838 mean intersection over union (mIoU) on their self-annotated dataset. 
They directly label ``lens artifacts'', a challenging source of error in shipborne ice data, where water droplets may land on the camera lens, partially or fully obscuring the environment in that region.
However, no suggestions were given on how to address and mitigate the loss of information in this portion of the image. 
Other works suggest creating approximations for the regions occluded by water droplets, through smoothing filters~\cite{icerainremoval} or neural networks~\cite{icegan, panchi2024weatherdegraded}. 
These approaches share an over-reliance on forms of spatial interpolation, which is a fundamental limitation.
Although it is possible to infer data in lightly occluded images or occluded images with simple geometry, the task of inferring complex geometries in heavily occluded regions is challenging. 
For example,  when looking at Figure~\ref{fig:unlabelrain}, it is almost impossible to infer what is blocked by the large water droplet without leveraging additional information or assumptions.

In this work, it is proposed to leverage temporal information to achieve more accurate predictions in occluded regions.
Predictions for the current frame are inferred from previous frames where the occlusion was not present at the same location as in the current frame.
To connect the current and previous frames when the ship is moving, methods based on optical flow are investigated.

\textbf{Optical Flow.}
Optical flow captures the motion of objects between frames by encoding pixel displacements as dense motion vectors. Early techniques relied on custom pixel encodings and linear transformations~\cite{FarnebackOpticalFlow, lucas1981opticalflow, horn1981opticalflow}, but recent advancements have favored neural networks for their superior ability to model complex motion patterns~\cite{ilg2016flownet2, sun2018pwcnet, yang2019hsmnet, teed2020raft}.
The present study employs PWCNet due to its use of warping and cost volume, which allows it to handle large displacements and occlusions.
By integrating warping and cost volume features from the PWCNet decoder, regions of invalid flow can be inferred.
This improves the network's ability to mitigate errors caused by occlusions, such as water droplets on the camera lens. 

\begin{figure}[t]
    \centering
    \adjustbox{width=.56\linewidth, center}{\parbox{\linewidth}{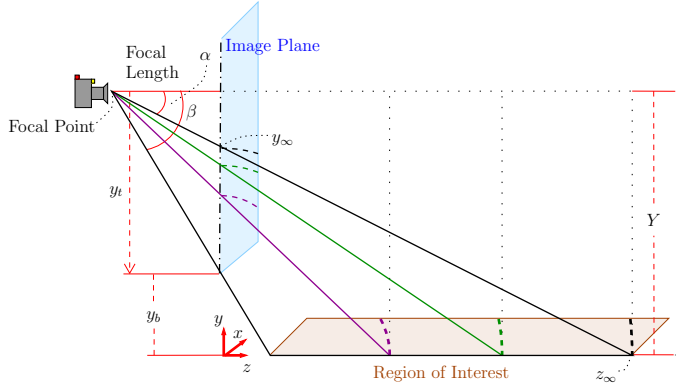}}
    \caption{Diagram demonstrating the relation between distance in the real-world and how it is projected onto the image plane. Equal distances in the region of interest are represented with dashed colored lines, which all converge to the focal point of the lens. The furthest point visible in the ROI is denoted as $z_\infty$, while the corresponding point in the image plane is denoted $y_\infty$. The camera is included for illustrative purposes.}
    \label{fig:regionDistIm}
\end{figure}

\textbf{Video semantic segmentation. }
Video semantic segmentation builds on semantic segmentation of image data by incorporating temporal information across sequential frames to enhance classification accuracy. This added dimension of temporal consistency is essential for tasks involving dynamic scenes, such as navigation in ice-covered waters. There are various techniques for extracting temporal information in video segmentation, including the use of recurrent networks~\cite{nilsson2017gru}, conditional random fields~\cite{3dcrf, chandra2018crf}, and, most notably, optical flow~\cite{gadde2017netwarp, ding2019efc, liu2020etc, huang2018videouncertainty}.

This work extends upon the approaches of Cheng et~al.~\cite{Chengsegflow} and Ding et~al.~\cite{ding2019efc}, who have integrated optical flow into video segmentation networks.
Cheng et al. proposed SegFlow, a network that combines ResNet~\cite{resnet} and FlowNet~\cite{fischer2015flownet} feature maps to jointly estimate optical flow and segment objects. 
However, their approach involves pooling and cropping the larger FlowNet features to match the size of the ResNet features, potentially losing valuable information, especially in lower-dimensional feature spaces. 
The present work addresses this limitation by ensuring features are only concatenated at the same level of the networks, ensuring the feature information is at similar scales and limiting the amount of lost information.
Additionally, Cheng et al.'s network is further modified by adjusting the ResNet encoder to have a six-channel input, and restructuring the decoder to resemble UPerNet, significantly improving segmentation performance without retraining on dedicated optical flow datasets.

Ding et al.~\cite{ding2019efc} sought to improve upon Cheng et al.'s design by incorporating occlusion prediction layers based on pixel similarity.
However, this heuristic-based approach can be limited by its inability to differentiate separate objects with similar colors.
Despite explicitly incorporating temporal knowledge into the network, their method does not outperform image segmentation networks that use similar techniques~\cite{zhu2019vplr}. 
In contrast, this work leverages PWCNet’s ability to infer temporal information and occluded regions dynamically, rather than relying on hard-coded pixel similarity.
By integrating the warping and cost volume information into a video segmentation network, superior performance to the tested image segmentation networks is achieved, particularly in challenging occluded regions, making the method presented more adaptable to real-world ice segmentation tasks.

\section{Dataset and Annotation} \label{dataset}

\subsection{Dataset Overview}
\begin{figure}[t]
    \centering
    \begin{minipage}[b]{.47\linewidth}
        \suboverpic[.85][magenta]{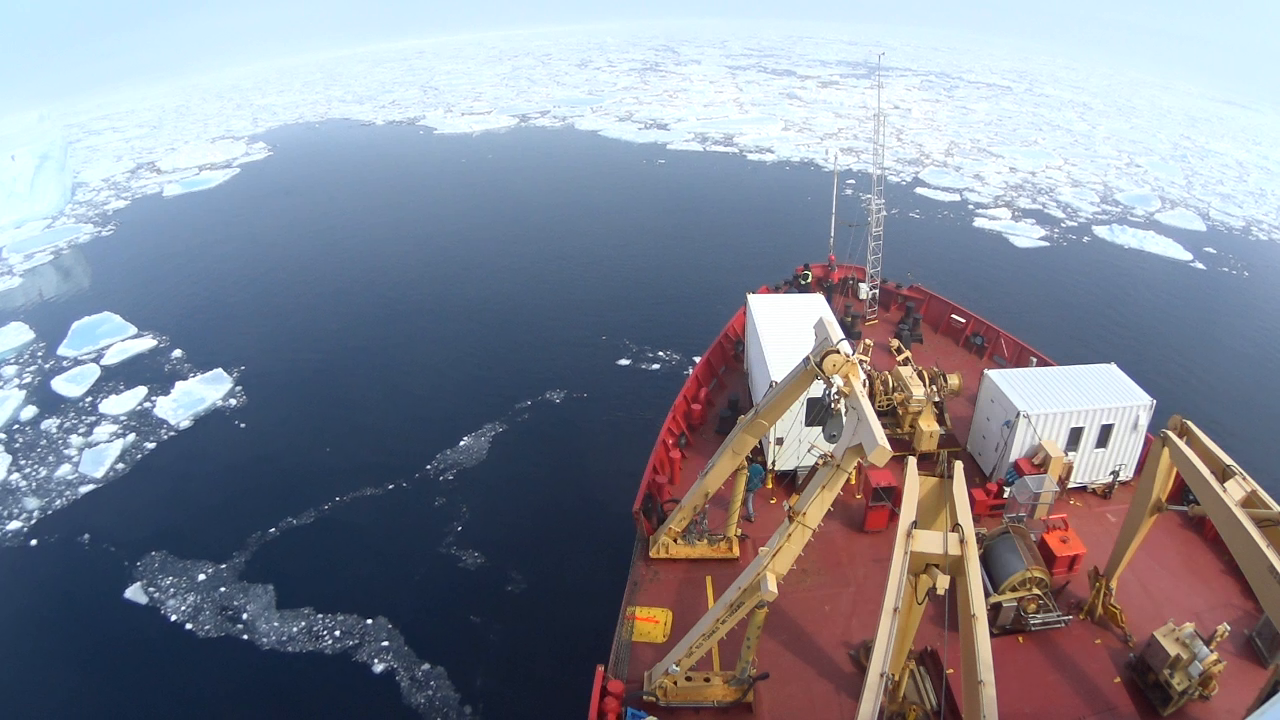}
    \end{minipage}
    \hfill
    \begin{minipage}[b]{.47\linewidth}
        \suboverpic[.85][magenta]{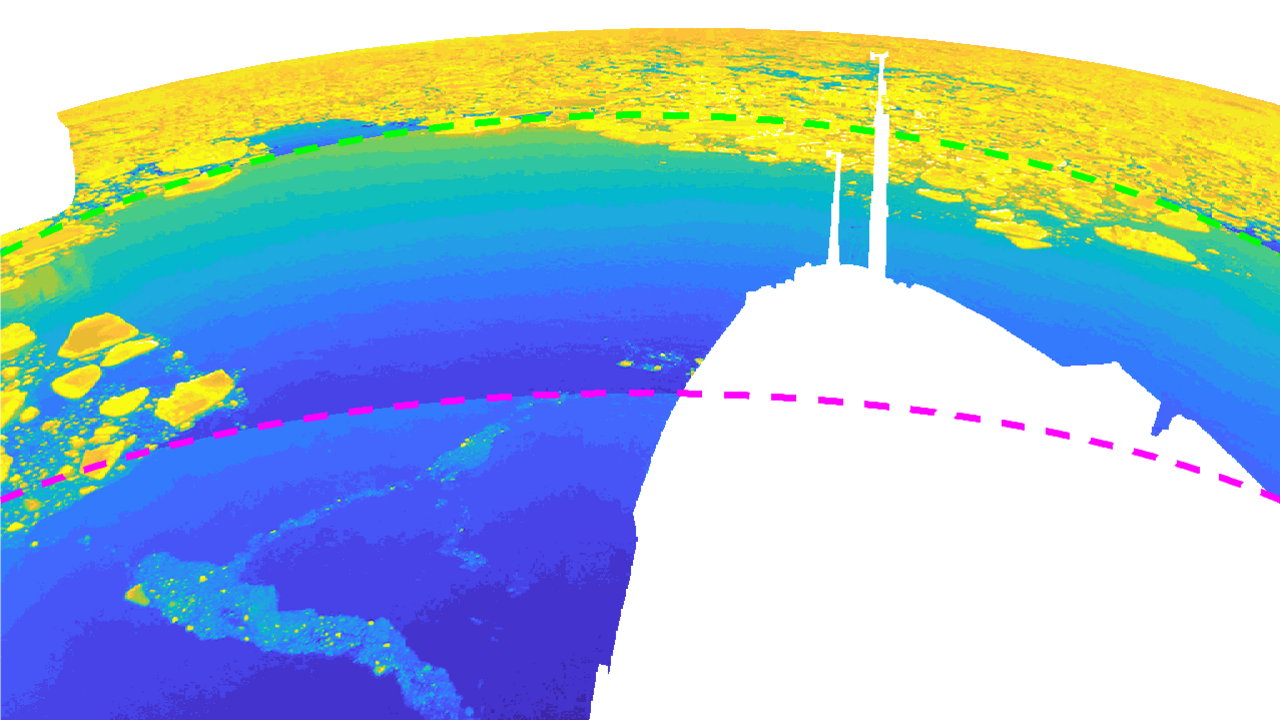} 
    \end{minipage}
    \\
    \begin{minipage}[b]{.47\linewidth}
        \suboverpic[.85][magenta]{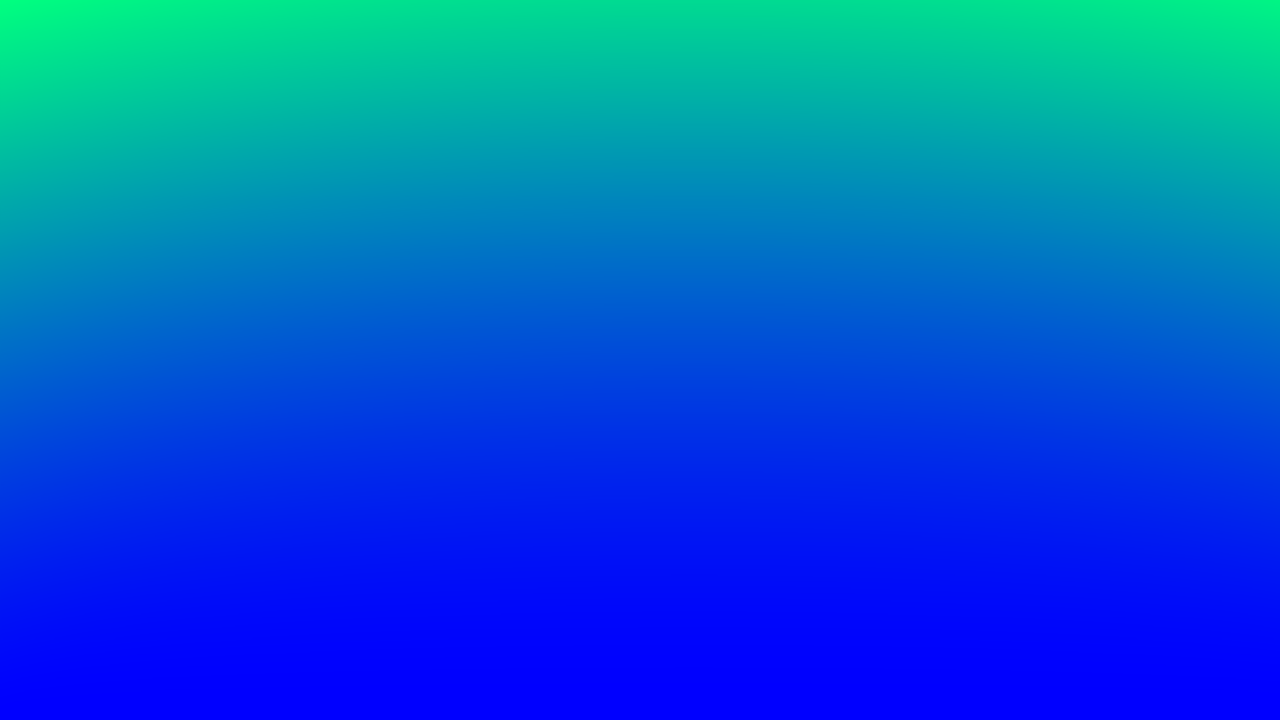}[waterthreshold]
        \hspace*{-4.5mm}
        \raisebox{-16mm}{
        \begin{Overpic}{
            \includegraphics[width=2mm, height=30mm]{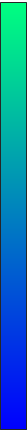}
        }
            \put(14,99){\scriptsize 340}
            \put(14,-3){\scriptsize 100}
        \end{Overpic} 
        }
    \end{minipage}
    \hfill
    \begin{minipage}[b]{.47\linewidth}
        \suboverpic[.85][magenta]{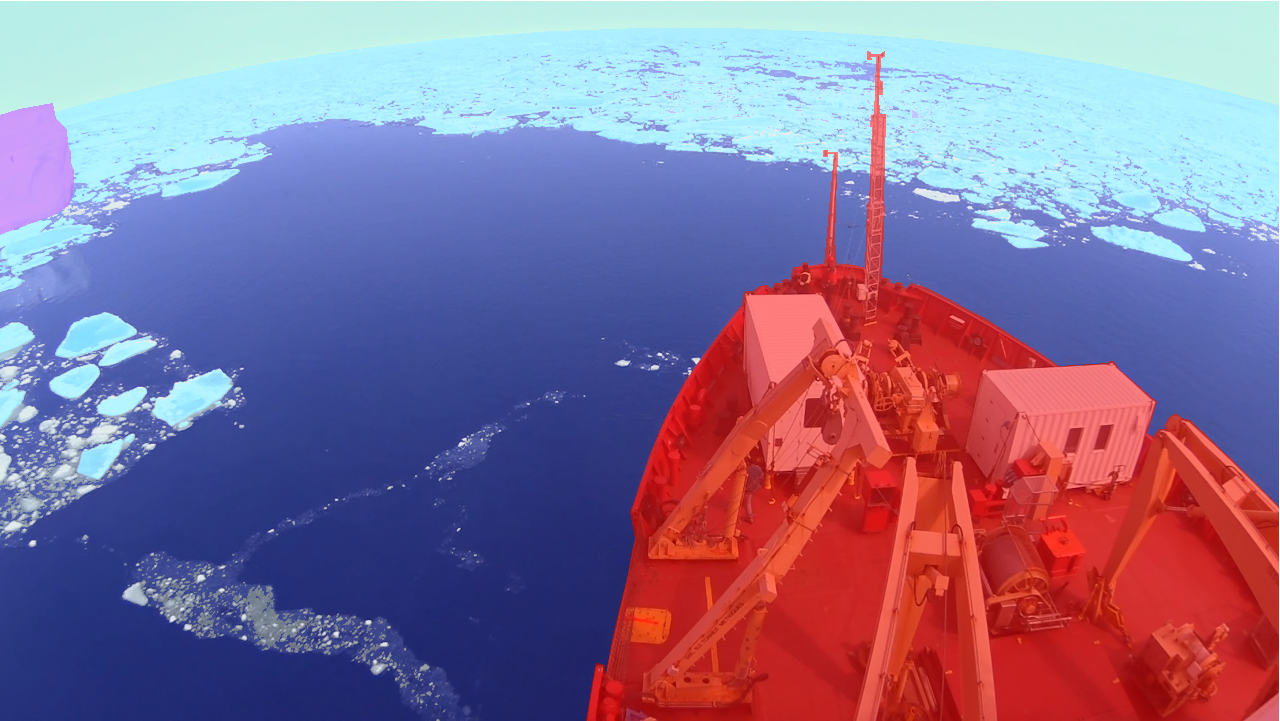}
    \end{minipage}
    \caption{The labeling process for the Amundsen dataset. \subref{fig:frame730} shows a manually labeled image, with the ship labeled in red, the sky in light green, and the iceberg in purple.  \subref{fig:frame730otsu} shows the result of MATLAB's \texttt{multithresh}, with the brightest classes in each region (represented by dashed multicolored lines) shown in yellow. \subref{fig:waterthreshold} shows the threshold distribution to label water, with pixels with intensity less than 100 labeled as water in the near-field. \subref{fig:frame730labeloverlay} shows the final annotation, overlaid on the original image.}
    \label{fig:labelingProcess}
\end{figure}
\begin{figure}
    \centering
    \begin{adjustbox}{max width=\linewidth, center}
        \suboverpic[.47][black]{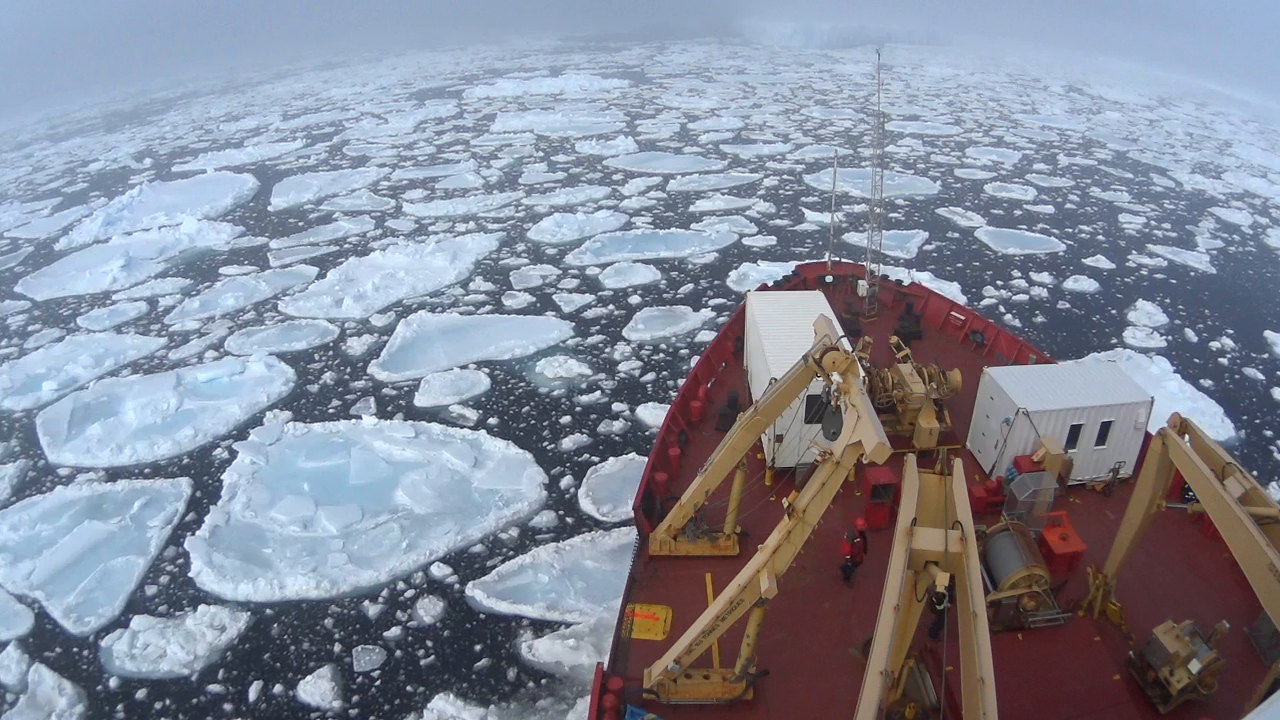} \hfill
        \suboverpic[.47][black]{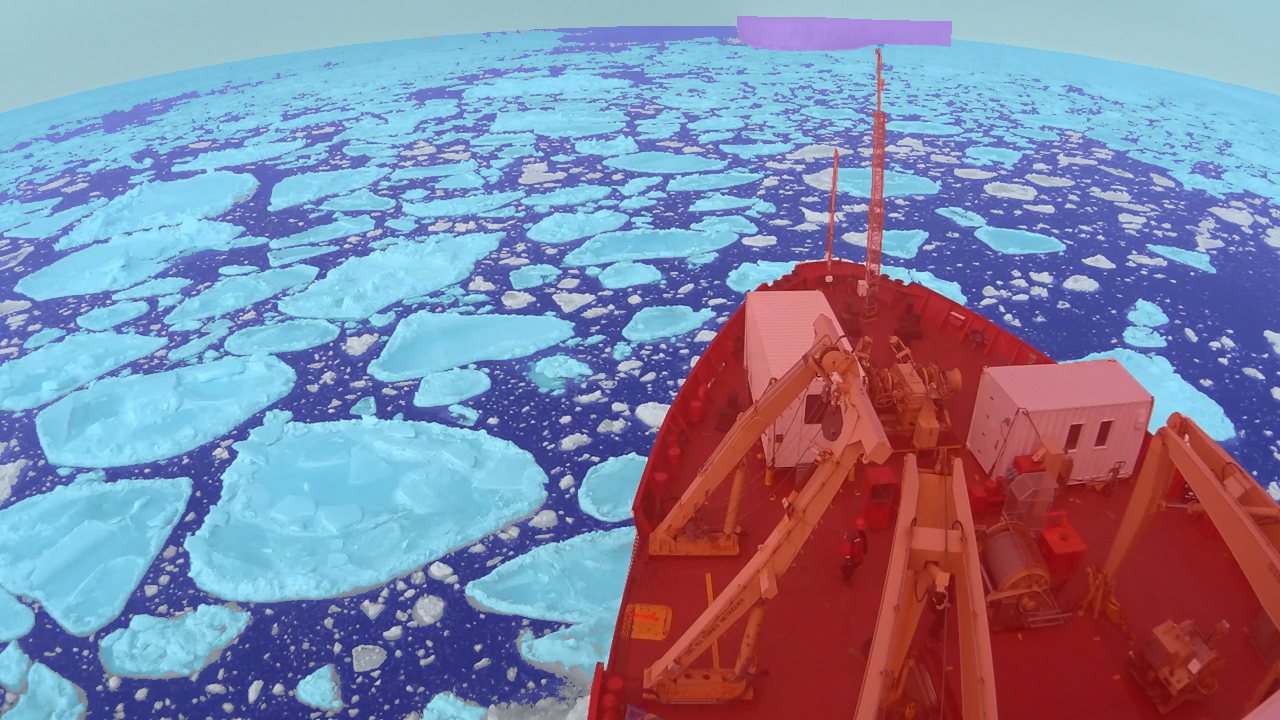}
    \end{adjustbox} \\
    \begin{adjustbox}{max width=\linewidth, center}
        \suboverpic[.47][black]{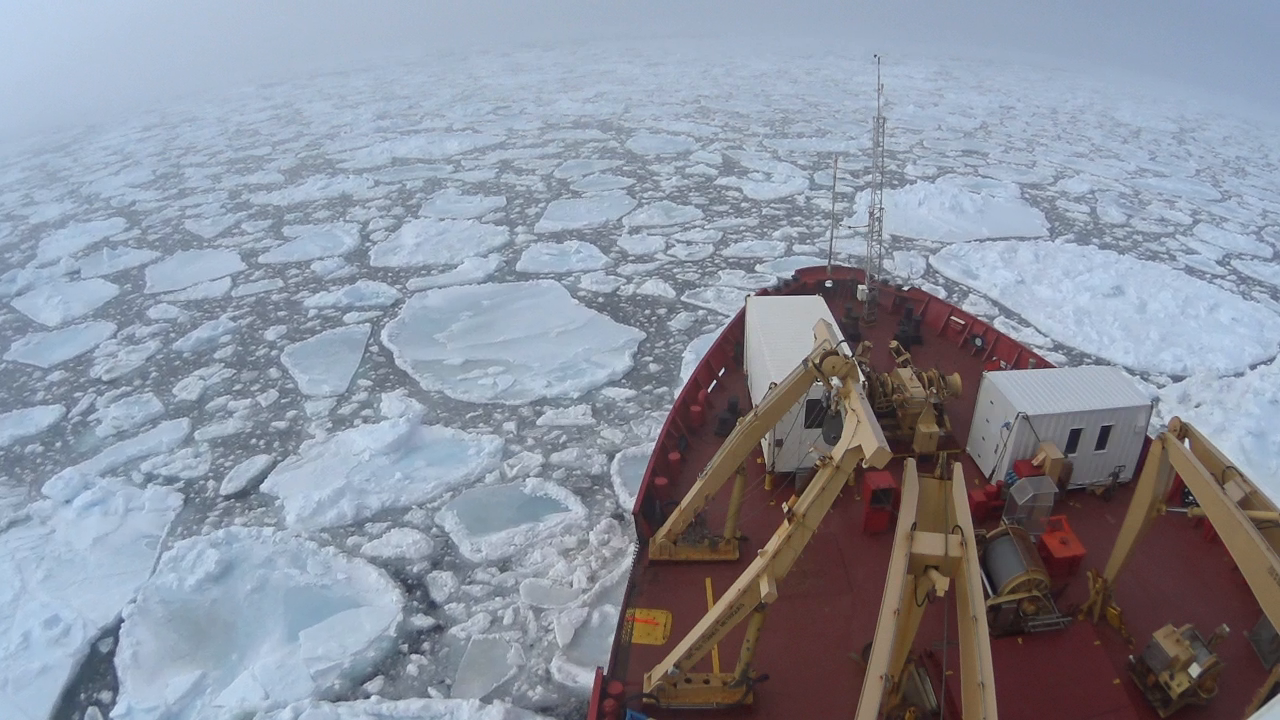} \hfill
        \suboverpic[.47][black]{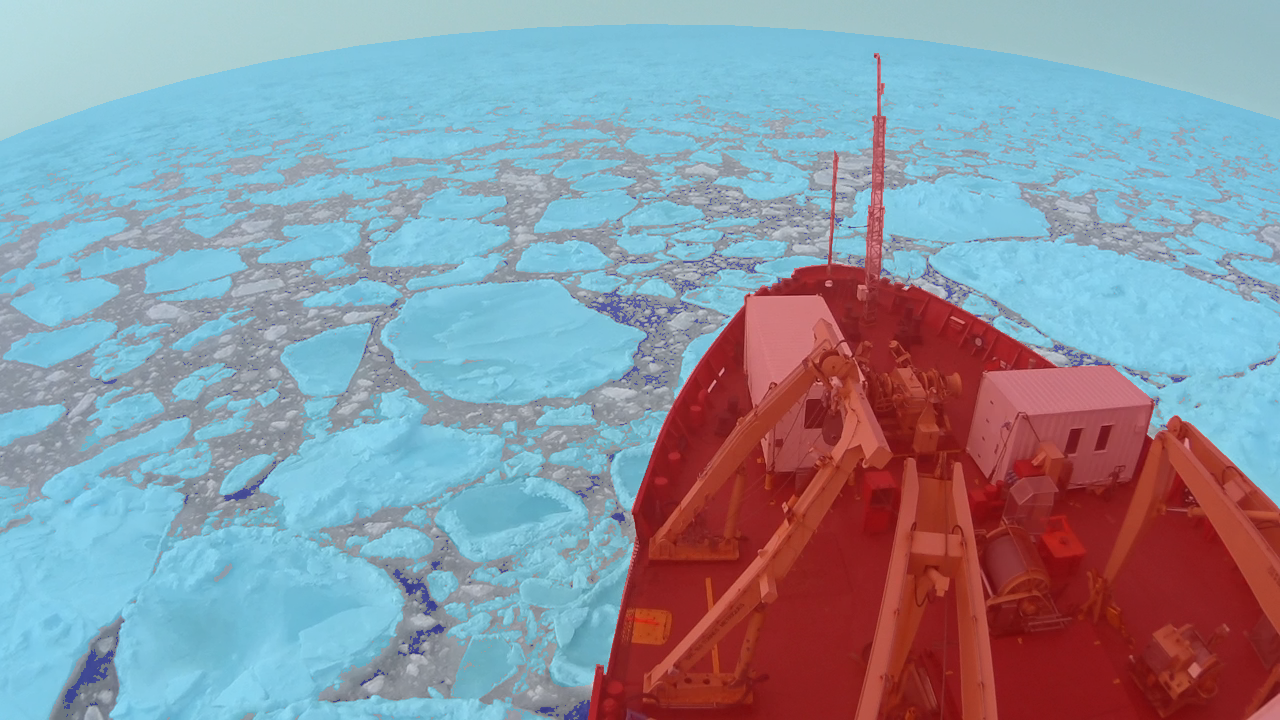}
    \end{adjustbox} \\
    \begin{adjustbox}{max width=\linewidth, center}
        \suboverpic[.47][black]{1_frame_360} \hfill
        \suboverpic[.47][black]{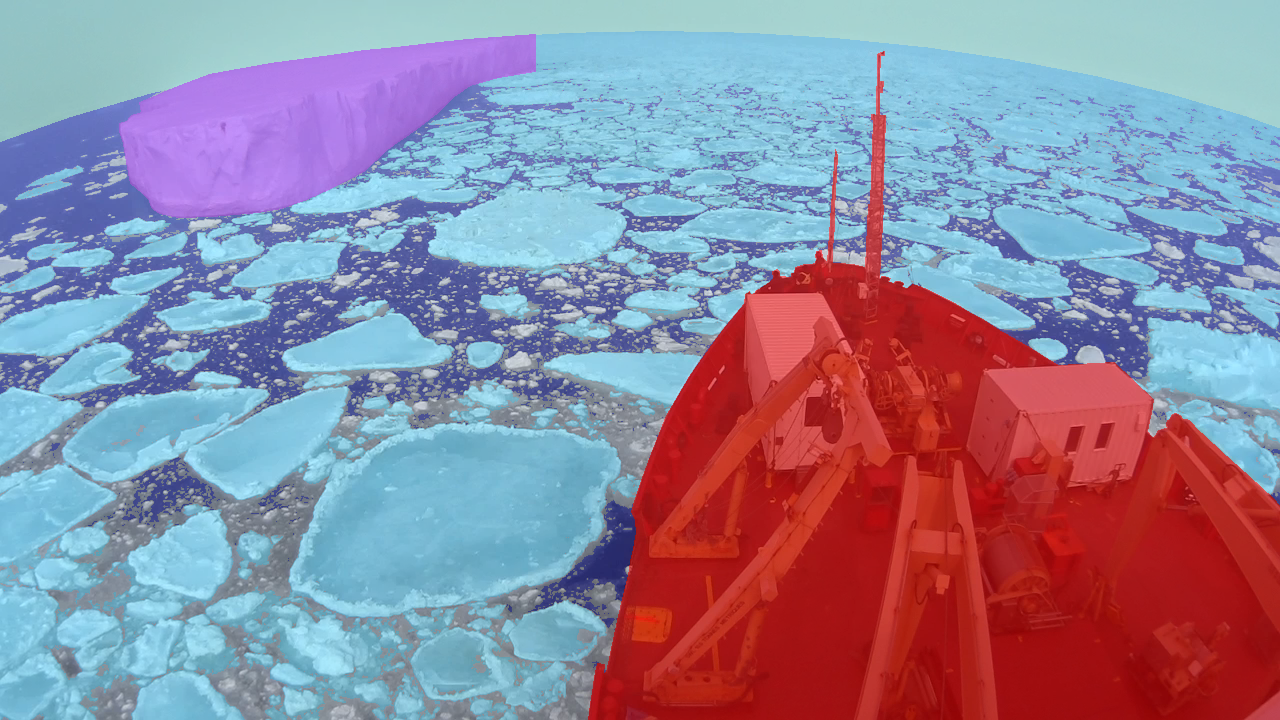}
    \end{adjustbox} \\
    \begin{adjustbox}{max width=\linewidth, center}
        \suboverpic[.47][black]{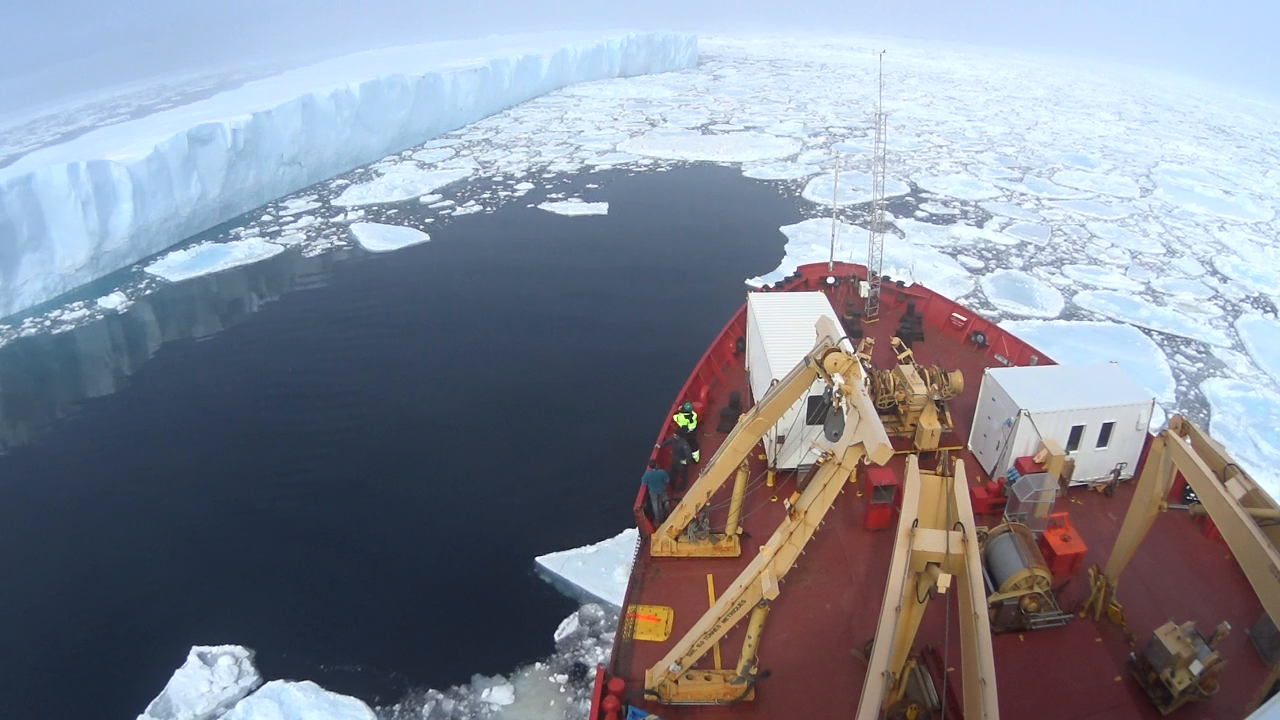} \hfill
        \suboverpic[.47][black]{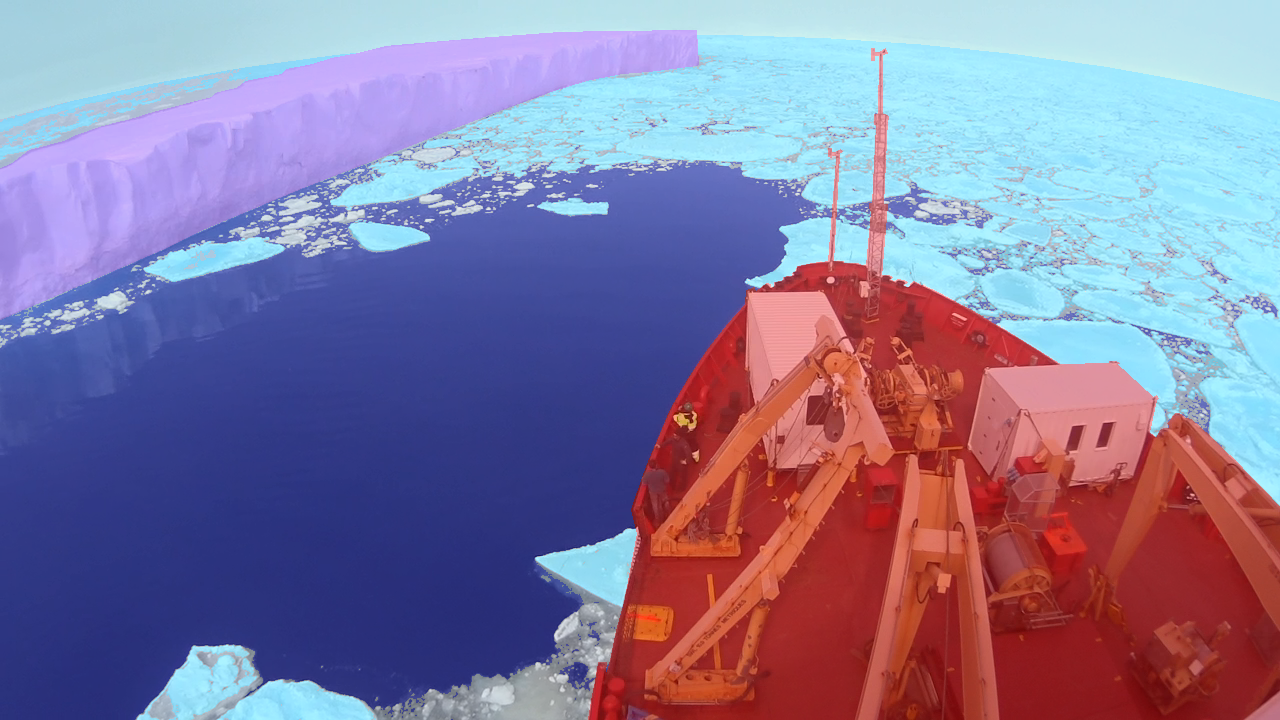}
    \end{adjustbox}
    \caption{Examples of labeled images from the Amundsen dataset. The original images are on the left, and the labels are on the right and overlaid on the original images for clarity. \subref{fig:1_frame_215} \& \subref{fig:1_frame_031} are examples from the training dataset. \subref{fig:1_frame_360} is an example from the validation dataset. \subref{fig:2_frame_050} is an example from the test dataset.}
    \label{fig:examplelabels}
\end{figure}

The dataset presented in this work follows the Canadian Coast Guard's icebreaker ``Amundsen'' during her expedition off the Newfoundland coast in the spring of 2015. 
The data feature varying conditions that range from ideal to messy due to factors such as lighting, lens obfuscations, and environmental conditions. Notably, this dataset includes footage of an iceberg, and to the authors' knowledge, is the first to provide and annotate in situ iceberg footage from onboard an icebreaker. 
The footage is featured in Jensen Huang's 2021 keynote~\cite{NVIDIAkeynote}, where it is highlighted as a new frontier application for machine learning.
Examples of the Amundsen dataset can be seen in Figures~\ref{fig:exampledataset}~and~\ref{fig:examplelabels}. 


The Amundsen dataset focuses on classifying and segmenting six distinct categories, following the World Meteorological Organization (WMO)~\cite{WMO}:
\begin{itemize}
    \item Iceberg: A massive piece of ice of varying shape, protruding more than 5 m above sea-level
    \item Ice floe: Any contiguous piece of sea ice
    \item Water: A large area of freely navigable water
    \item Brash ice: Weakly frozen together clusters of ice crystals and small fragments of floating ice no more than 2~m across
    \item Ship: The boat present in the near-field\footnote{The term ``near-field'' is distinct from the term ``close-range'' in that the near-field refers to the immediate area closest to the camera, while close-range encompasses a broader region of the captured imagery, including but not limited to the near-field.} of all images; including the artifacts and people on board
    \item Sky: All visible sky in the image
\end{itemize}

\subsection{Region-Based Annotation}
Many works leverage distinct features of the ice in order to segment it; including shape, texture, and color~\cite{zhang2014otsukmeans, kalke2018svm}. This work postulates that color is the most important differentiating feature of ice and water from their surroundings, and color-thresholding techniques can be used sufficiently when combined with heuristics. In this section, color-based algorithms are applied to distinct regions in the image based on their distance and lighting; a methodology dubbed Region-Based Annotation.

\subsubsection{Defining Regions}

The extrinsic properties of the camera define the transformation between the real-world and the image plane as a result of perspective, while the intrinsic properties describe the transformation caused by the physical properties of the lens~\cite{cameraCalibration}.
These properties are required to determine the size and distance of the features in an image. However, they are not provided for the Amundsen dataset.
Thus, a rough parametric model is constructed to describe the extrinsic and intrinsic properties of the data.

\textbf{Extrinsic Properties Model.}
Figure~\ref{fig:regionDistIm} depicts the expected relationship between the real-world and the image plane.
The focal point, situated within the camera lens, serves as the convergence point for incoming light rays.
The image plane is where light is projected on to form the image.
The distance between the focal point and the image plane is defined as the focal length ($f$).
Within view of the camera lies the region of interest (ROI), which represents the landscape being captured.
The furthest point visible in the ROI is denoted as $z_\infty$, while the corresponding point in the image plane is denoted $y_\infty$.

Let $y_t$ represent the vertical distance from the focal point in the image plane, $Y$ represent the ROI's vertical distance from the camera, and $z$ represent the ROI's horizontal distance from the camera.
Using similar triangles, the relationship between $y_t$ and $z$ can be expressed as, $
    \frac{y_t}{Y}=\frac{f}{z+f}$.
This expression can be redefined to originate from the bottom of the image, then isolated for the image coordinate $y$, becoming,
\begin{equation}
    y=Y\left(\gamma-\frac{f}{z+f}\right),
    \label{eq:extrinsicRelation}
\end{equation}
where $\gamma = 1-\frac{y_b}{Y}$, and $y_b\in[0,Y]$ is the vertical distance from the bottom of the image plane to the region of interest.
$\gamma \in [0,1]$ can be thought of as the ratio between $Y$ and the vertical field of view of the camera.

This relation can be simplified further through normalization.
Let $z_0$ and $z_\infty$ denote the closest and furthest points in the ROI, respectively, with corresponding image plane points $y_0$ and $y_\infty$.
These points are normalized such that the coordinates of the closest and furthest points in both the ROI and the image plane range between 0 and 1.
The normalized coordinates are defined as $y^* = \frac{y}{y_\infty}$ and $z^* = \frac{z}{z_\infty}$, while $Y^* = \frac{Y}{y_\infty}$ and $f^* = \frac{f}{z_\infty}$ are the normalized versions of the original parameters $Y$ and $f$, respectively. By constraining the normalized coordinates to $y^*|_{z^*=0} = 0$, and $y^*|_{z^*=1} = 1$ into Equation~\eqref{eq:extrinsicRelation}, it can then be found that $\gamma = 1$ and $f^* = Y^*-1$.

The final simplified relation is given by,
\begin{equation}
    y^*=Y^*\left( \frac{z^*}{z^*+Y^*-1}\right).
    \label{eq:normalizedExtrinsic}
\end{equation}
This mimics the physical extrinsic model given by Equation~\eqref{eq:extrinsicRelation} and depends only on a single parameter, $Y^*$.

\textbf{Intrinsic Properties Model.}
The images are also expected to be physically warped from the physical properties of the lens, known as the intrinsic properties of the camera. To account for this, it is assumed that the transformation for the lens's intrinsic properties can be mostly described with a linear scaling operation. 
Distance to the camera can then be represented by an elliptical coordinate system, which can be thought of as a skewed polar coordinate system.
Cartesian coordinates can be converted to elliptical coordinates using the relation 
\begin{equation}
    \mu = \frac{\acosh{ \left(\sqrt{(x+a)^2 + y^2} + \sqrt{(x-a)^2 + y^2}\right)}}{2a}.
    \label{eq:intrinsicrelation}
\end{equation}
where $\mu$ represents the elliptical curves, and $a$ represents the location of the ellipse foci~~\cite{coordinateSystem}.
The distances between the ellipses are modeled with the extrinsic property relation in Equation~\eqref{eq:extrinsicRelation}.

\subsubsection{Semi-Manual Annotation} \label{semimanualannotation}
The optical images are labeled sequentially in a multi-pass approach. The ship, sky, and iceberg are labeled manually, since each are large, easily identifiable objects, that additionally contain colors that are hard to distinguish from the rest of the image. To reduce manual annotation costs, the video format of the data is leveraged by creating sparse annotations over time and interpolating between these frames. The dataset is annotated with CVAT (Computer Vision Annotation Tool)~\cite{cvat}, labeling every 10th image while manually correcting minor errors that arise from interpolation. 
Figure~\ref{fig:frame730} shows an example of the manual annotation (with corresponding color) of the ship (red), sky (light green), and iceberg (purple).

After the large objects are classified, the ice floes are annotated.
The image is separated into multiple regions based on their distance from the camera, and oversegmented using MATLAB's \texttt{multithresh} function, a multiclass extension of Otsu's algorithm.
The algorithm is used to generate multiple proposals for the color threshold of ice floes, significantly reducing the effort required to determine the optimal threshold.
In each region, the threshold is chosen such that the brightest classes are selected as ice floes.
Figure~\ref{fig:frame730otsu} shows a typical result, with the brightest classes in each region shown in yellow.
The ice floe classification is finalized by removing small clusters using MATLAB's \texttt{bwareafilt}, to avoid misclassifying small ice fragments.

To annotate the brash ice and water, a distribution of blue color thresholds is interpolated over the coordinate system described by Equations~\eqref{eq:normalizedExtrinsic}~and~\eqref{eq:intrinsicrelation}.
A linear model is chosen for simplicity, such that a continuous range of thresholds can be created using only two values.
Figure~\ref{fig:waterthreshold} shows the threshold distribution for the given image, and Figure~\ref{fig:frame730labeloverlay} shows the final annotation overlaid on the raw image.

Additional details for the labeling process can be found in \ref{labelingdetails}, and specific values for the labeling hyperparameters can be found in Table~\ref{tab:labelingvalues}.

Figure~\ref{fig:examplelabels} shows examples of the labels of the Amundsen dataset. This approach was used to annotate 945 labeled images, from 3 separate videos based on their relatively clean data and balanced distribution of classes. The dataset was split approximately 80:10:10 for training, validation, and testing, respectively, with two videos used for training and validation and one video being held out for testing.
The annotated Amundsen dataset provides precise descriptions of the surrounding ice environment and enables large neural networks to learn effectively from the data.

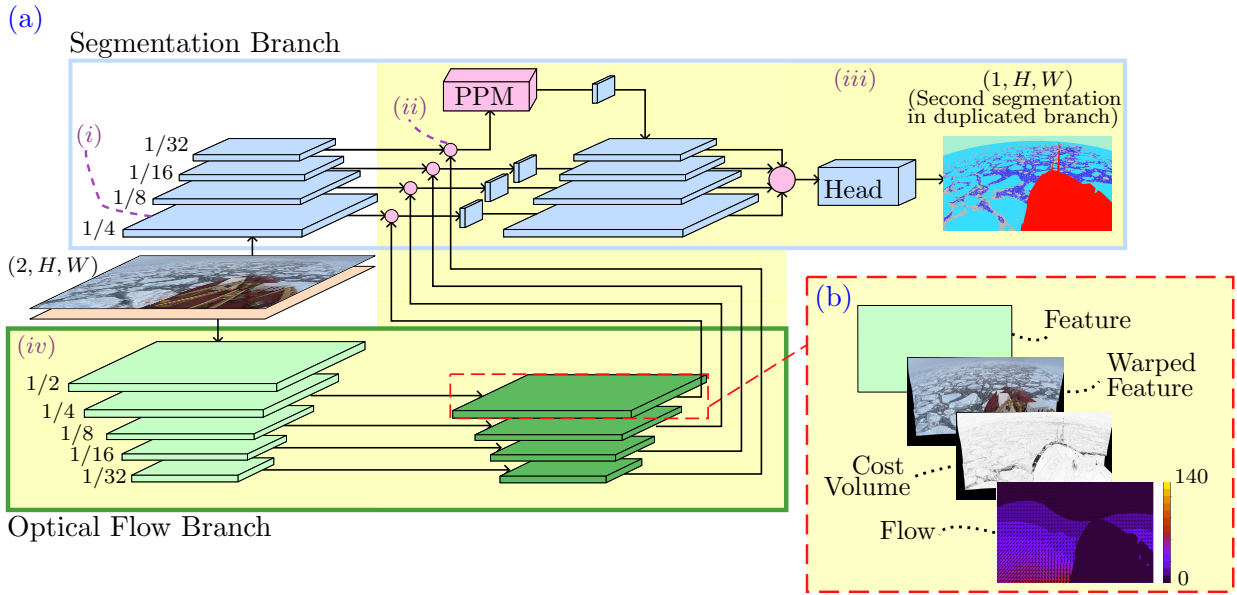
\begin{figure*}[h!]
    \centering
    \begin{subcaptiongroup}
        \begin{overpic}[width=\textwidth]{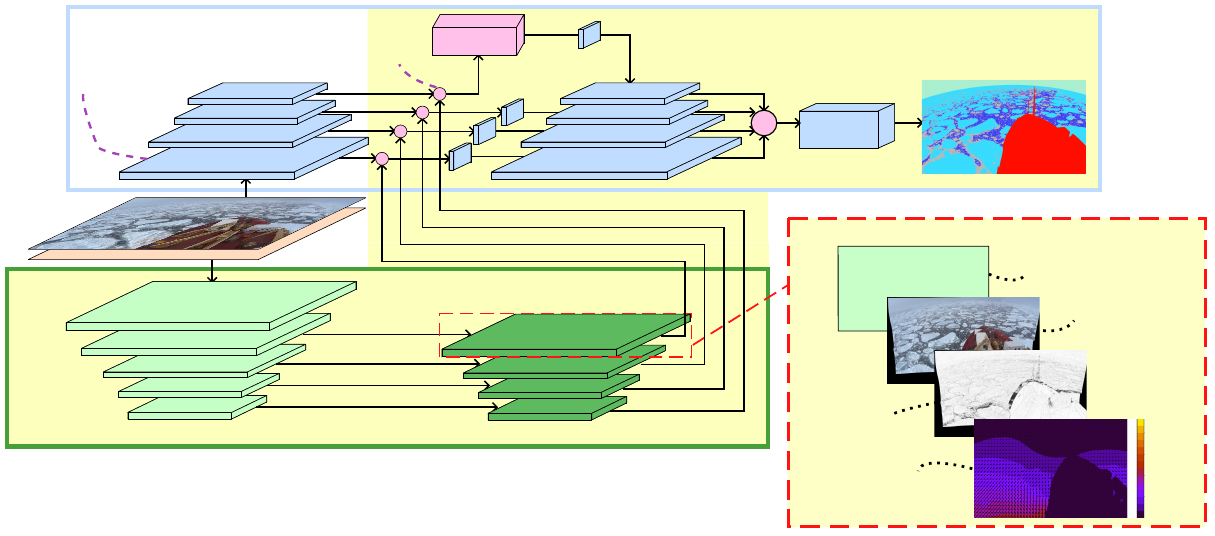}
        \put(36.5, 40){\small PPM}
        \put(66.25, 32.5){\small Head}
        \put(5.5, 44.25){Segmentation Branch}
        \put(0.5, 5.25){Optical Flow Branch}
        \put(84, 22){\small Feature}
        \put(89, 18.5){\small{Warped}}
        \put(89, 16.5){\small{Feature}}
        \put(68.5, 10.25){\small{Cost}}
        \put(66.25, 8.5){\small{Volume}}
        \put(70.75, 5){\small{Flow}}
        \put(94.5, 9.5){\scriptsize 140}
        \put(94.75, 1.25){\scriptsize 0}
        \put(0.5, 26.5){\scriptsize{$(2, H, W)$}}
        \put(6.5, 29.5){\scriptsize{$1/4$}}        
        \put(9, 32){\scriptsize{$1/8$}}
        \put(10.25, 34){\scriptsize{$1/16$}}
        \put(11.5, 36.25){\scriptsize{$1/32$}}
        \put(2, 17){\scriptsize{$1/2$}}
        \put(3.25, 14.75){\scriptsize{$1/4$}}
        \put(4.75, 13){\scriptsize{$1/8$}}
        \put(5.5, 11.25){\scriptsize{$1/16$}}
        \put(6.5, 9.5){\scriptsize{$1/32$}}
        \put(79, 41.5){\scriptsize{$(1,H,W)$}}
        \put(73, 40){\scriptsize{(Second segmentation}}
        \put(73, 38.5){\scriptsize{in duplicated branch)}}
        \put(6, 37){\footnotesize{\color{Fuchsia}{\textbf{$(i)$}}}}
        \put(31.5, 39.25){\footnotesize{\color{Fuchsia}{\textbf{$(ii)$}}}}
        \put(67, 41.5){\footnotesize{\color{Fuchsia}{\textbf{$(iii)$}}}}
        \put(1.25, 20){\footnotesize{\color{Fuchsia}{\textbf{$(iv)$}}}}
        \captionlistentry{UPerFlow Architecture}
        \put(0.5, 46.25){\textcolor{blue}{(\thesubfigure)} \label{fig:uperflow_components}}
        \captionlistentry{PWCNet Snapshot}
        \put(65.5, 23.75){\textcolor{blue}{(\thesubfigure)} \label{fig:pwcnet_components}}
    \end{overpic}
    \end{subcaptiongroup}
    \caption{The illustrated framework for UPerFlow is depicted in~\subref{fig:uperflow_components}, where UPerNet serves as the segmentation branch (blue) and PWCNet serves as the optical flow branch (green). Concatenation operations are represented by pink circles, and the Pyramid Parsing Module of PSPNet~\cite{zhao2017pspnet} is denoted by PPM. The optical flow branch is depicted predicting the backward flow, while the segmentation branch generates segmentations for the first image. Yellow-highlighted sections indicate duplicated components within the architecture, responsible for predicting forward flow and segmentations for the second image. The contributions of this work are listed in purple as follows: \textcolor{Fuchsia}{(i)} a six-channel input ResNet encoder, \textcolor{Fuchsia}{(ii)} cross-connections from PWCNet, \textcolor{Fuchsia}{(iii)} duplicated segmentation decoder, one for each image, and \textcolor{Fuchsia}{(iv)} duplicated optical flow branches to predict bi-directional flow. \subref{fig:pwcnet_components} provides a detailed view of the decoder blocks in PWCNet, showing the integration of optical flow, cost volume, and warped features at the low-level feature stage.}
    \label{fig:uperflow_diagram}
\end{figure*}

\section{Classification and Segmentation Methodology} \label{methodology}

This section presents UPerFlow, an optical flow-based segmentation network.
The network is evaluated against the top-performing image segmentation network from a suite of six CNN's, UPerNet with a 101-layer ResNet backbone.
The video architecture of UPerFlow allows it to outperform some of the best-performing networks used in previous studies (Section~\ref{background}), especially in occluded regions.

\subsection{Video Semantic Segmentation - UPerFlow}
To improve segmentation in occluded areas, UPerFlow combines optical flow and segmentation into a unified network.
The architecture includes a segmentation encoder branch, an optical flow encoder branch, and segmentation decoders with skip connections. 
In the decoders, features from the final four layers of each branch are concatenated to the decoders with cross connections to generate segmentation predictions.
The architecture follows a modified version of SegFlow~\cite{Chengsegflow}. The architecture is illustrated in Figure~\ref{fig:uperflow_diagram}.

\textbf{Segmentation Encoder Branch.}
A pretrained 101-layer ResNet encoder is used for segmentation, modified to accept six channels to process RGB channels from both images. This six-channel approach is consistent with techniques in optical flow models like FlowNet~\cite{fischer2015flownet, ilg2016flownet2} and RAFT~\cite{teed2020raft}.

\textbf{Segmentation Decoder.}
Two separate decoders generate segmentations, one for for each image.
Each are based on the architecture of UPerNet~\cite{xiao2018upernet}, since it was found to be the best-performing of the tested image segmentation networks. However, the framework can adapt to alternative architectures as needed. The duplicated components of UPerFlow are highlighted in yellow in Figure~\ref{fig:uperflow_components}.

\begin{figure}[t]
    \centering
    \adjustbox{max width=\linewidth, center}{%
         \suboverpic[.3][green][2,88]{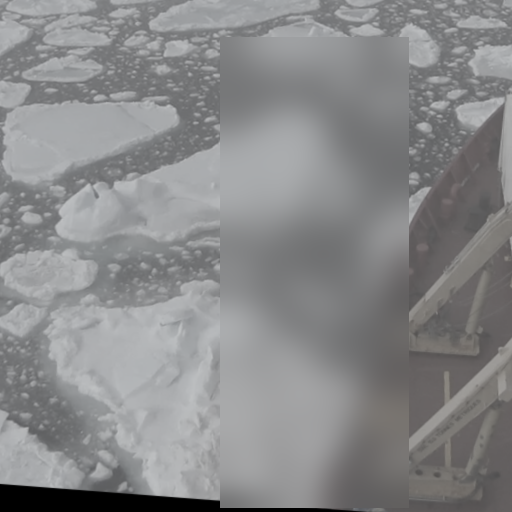}[minAugmentation]%
         \suboverpic[.3][black][2,88]{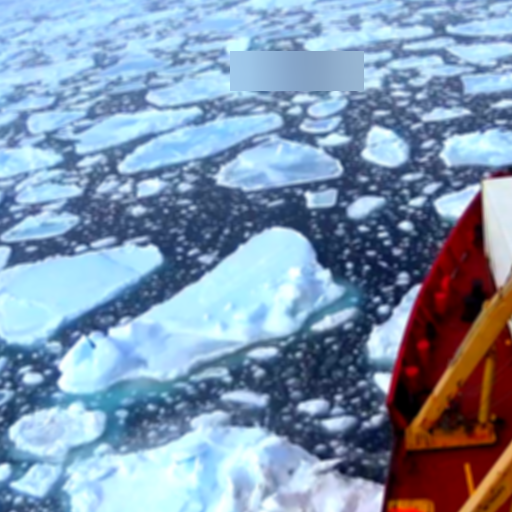}[maxAugmentation]%
         }
    \caption{Extreme examples of data augmentations from the data augmentation pipeline. \subref{fig:minAugmentation} shows the randomly cropped image with low color jittering, rotation, gaussian blur, and random erasing. \subref{fig:maxAugmentation} shows the opposite end of the spectrum from \subref{fig:minAugmentation}.}
    \label{fig:dataaugmentation}
\end{figure}

\textbf{Optical Flow Branch.}
PWCNet is selected for its robust optical flow capabilities and straightforward CNN design. It uses two encoders to process each image separately before merging into a unified optical flow decoder. In a UNet-like structure, PWCNet includes skip connections between encoder and decoder layers, alongside warping and cost volume features. The warp layer transforms encoded features from one image using the optical flow prediction in the decoder, and the cost volume layer ($CV$) compares the warped features with the original image features through a correlation
\begin{equation}
    CV_{i,j}(l_1, l_2) = \frac{1}{N} (c_i(l_1))^T c_j(l_2),
\end{equation}
where $N$ is the number of channels in the $i$-th and $j$-th column vectors of latent features $l_1$ and $l_2$, denoted as $c_i(l_1)$ and $c_j(l_2)$, respectively. The components of PWCNet are illustrated in Figure~\ref{fig:pwcnet_components}.

\textbf{Cross-connections.}
To incorporate temporal information into segmentation predictions, optical flow features are added to the segmentation decoders. 
Specifically, the final four layers of PWCNet, including flow, warp, and cost volume features, are concatenated with corresponding ResNet features from the segmentation encoder at matching length scales, thus preserving information in the latent space without cropping. 
The approach can be thought of as a modified UPerNet architecture, where each cross-connection incorporates additional optical flow information.

\textbf{Bi-directional Flow.}
Bi-directional optical flow captures temporal context by generating forward flow $f_{n\rightarrow (n+1)}$, representing motion from the first to the second image, and backward flow $f_{(n+1)\rightarrow n}$ representing the reverse.
Each image pair is passed through PWCNet twice, in both chronological and reverse order, to extract both directions of flow.
Forward flow is integrated into the segmentation decoder for the second image, while backward flow provides context for the first.
This enhances accuracy in cases of occlusion present in one image, where the flow, warp, and cost volume features from the other image may provide context for the otherwise missing information in the occluded region.

\subsection{Training and Evaluation} \label{semseg}
A large Bayesian hyperparameter search using Optuna~\cite{optuna_2019} was conducted to simultaneously optimize the choice of image segmentation network, along with training and data augmentation hyperparameters. UPerFlow was then trained using the same hyperparameters as the top-performing model, UPerNet.

Additional implementation details regarding the model can be found in \ref{modeldetails}, with the training hyperparameter values shown in Table~\ref{tab:traininghpvalues}, and the data augmentation values in Table~\ref{tab:dataaugmentationvalues}.

A notable technique used in the training of the networks is random erasing~\cite{zhong2017randomerasing}, where rectangular regions of random proportion are erased in an image.
These regions are replaced with either the mean value of the image or a random Gaussian blur.
This technique is used to simulate occlusions, where a raindrop on the lens can either partially or fully obscure a portion of the image.
The power level of the blur is chosen randomly, with the standard deviation of the Gaussian kernel chosen to be within 15 to 100 pixels, representing different strengths of occlusion.
UPerFlow is trained with asymmetric random erasing, where only one image from the pair can have random erasing.
Figure~\ref{fig:dataaugmentation} shows examples of the data augmentation pipeline, and different strengths of occlusion. 

The networks are evaluated using the mean Intersection over Union (mIoU) and mean pixel-wise accuracy (mAcc) on the test dataset. Additionally, since the test dataset does not include natural occlusion, two new datasets are constructed by artificially augmenting the test dataset with random erasing. The artificial datasets are created by randomly erasing square portions of every other image with a Gaussian blur.
These square portions can either represent light occlusions using a Gaussian blur standard deviation of 15 pixels or strong occlusion with 100 pixels.
The kernel size is fixed and occludes 15\% of the total image area.
This approach emulates the technique used in training and allows the networks to demonstrate the extent of their predictive capabilities for familiar occlusions.

\section{Results and Discussion} \label{results}
\subsection{Occlusion-Free Performance} \label{semsegresults}
\begin{table*}[t]
    \centering
    \setlength{\tabcolsep}{0.5em}
    \begin{tabular}{|C{3cm} |C{2cm}  |C{2cm}|C{2cm}|C{2cm}|C{2cm}|}
        \hline
        \multirow{2}{3cm}{\centering\textbf{Network Architecture}} & \multirow{2}{2cm}{\centering\textbf{Occlusion Level}} &\multicolumn{2}{M{4.2cm}|}{\textbf{Overall Performance}}& \multicolumn{2}{M{4.2cm}|}{\textbf{Occluded Region Performance}}\\
        \cline{3-6}
         & & \textbf{mIoU} & \textbf{mAcc} & \textbf{mIoU} & \textbf{mAcc} \\
        \hline
        \multirow{3}{*}{UPerNet} & None & 0.838 & 0.946 & (-) & (-) \\
         & Light & 0.829 & 0.941 & 0.657 & 0.892 \\
         & Heavy & 0.819 & 0.935 & 0.535 & 0.818 \\
        \hline
        \multirow{3}{*}{UPerFlow} & None & \textbf{0.844} & \textbf{0.948} & (-) & (-) \\
         & Light & \textbf{0.837} & \textbf{0.945} & \textbf{0.727} & \textbf{0.911} \\
         & Heavy & \textbf{0.835} & \textbf{0.943} & \textbf{0.736} & \textbf{0.897} \\
        \hline
    \end{tabular}
    \caption{Performance metrics of UPerFlow and UPerNet across varying occlusion levels. Overall performance is shown alongside performance measured exclusively in the occluded regions. UPerFlow demonstrates improved performance over the baseline in all conditions, with substantial gains in occluded areas.
    }
    \label{tab:video_occlusion_results}
\end{table*}

 The performance metrics on the occlusion free dataset are shown in Table~\ref{tab:video_occlusion_results}, denoted by ``None''. On the test dataset, UPerNet101 achieved an mIoU of 0.838 and an mAcc of 0.946. UPerFlow outperforms this baseline performance with an mIoU of 0.844 and mAcc of 0.948, corresponding to a half percent increase in mIoU and a marginal increase in mAcc. Embedding the flow features from PWCNet directly into the network provides UPerNet with temporal context, resulting in improved performance overall.

Inference results on occlusion-free data are shown in Figure~\ref{fig:test_results}, denoted ``None''.
The performance of both models are similar, each maintaining a strong ability to segment the various classes within the dataset. 
The improvement of UPerFlow is provided in certain `corner cases' of the test dataset.
For example, UPerFlow demonstrates an apparent reduction in misclassifications caused by the reflection of the iceberg in the water.
Additionally, it better classifies the region above the iceberg, preferring to identify the region as brash ice, rather than erroneously continuing the iceberg prediction to the horizon.
This suggests that the optical flow network’s understanding of coherent objects plays a critical role in aiding the segmentation process.
By recognizing the iceberg as a distinct object, the optical flow network likely helps the segmentation network generate more accurate predictions in these challenging areas. 

However, the model still exhibits some errors. 
Due to inaccuracies in optical flow, the network tends to generate erroneous predictions near image and object boundaries.
For instance, in Figure~\ref{fig:test03uperflow}, the network misclassifies the ice floe adjacent to the ship. 
Similarly, Figure~\ref{fig:test43uperflow} illustrates how the network misclassifies the iceberg at the edge of the image. 
However, these errors are rare, and they are relatively minor compared to the network's improved ability to handle corner cases and occlusions. 

\subsection{Occlusion Mitigation} \label{semsegocc}

\newcommand{\unc}{\cellcolor{red!20}}
\newcommand{\ufc}{\cellcolor{red!20!green!10}}

\begin{sidewaysfigure*}

\addtolength{\tabcolsep}{-0.4em}
\begin{tabular}[c]{>{\centering\arraybackslash}p{0.24\linewidth}c>{\centering\arraybackslash}p{0.24\linewidth}>{\centering\arraybackslash}p{0.24\linewidth}>{\centering\arraybackslash}p{0.24\linewidth}}
    \rowcolor{blue!10}\
    \textbf{Ground Truth} & & \textbf{None} & \textbf{Light} & \textbf{Heavy} \\%
    \rowcolor{blue!5} & \unc \rotcentered{UPerNet} &
    \unc\begin{minipage}[c][3cm]{\linewidth}%
        \centering
        \suboverpic{test03upernet}%
    \end{minipage} &%
    \unc\begin{minipage}[c][3cm]{\linewidth}%
        \centering
        \suboverpic{occlusionlight03upernet}%
    \end{minipage} &%
    \unc\begin{minipage}[c][3cm]{\linewidth}%
        \centering
        \suboverpic{occlusionheavy03upernet}%
    \end{minipage}\\%
    \rowcolor{blue!5}\multirow{-8}{*}{\includegraphics[width=.9\linewidth, valign=m]{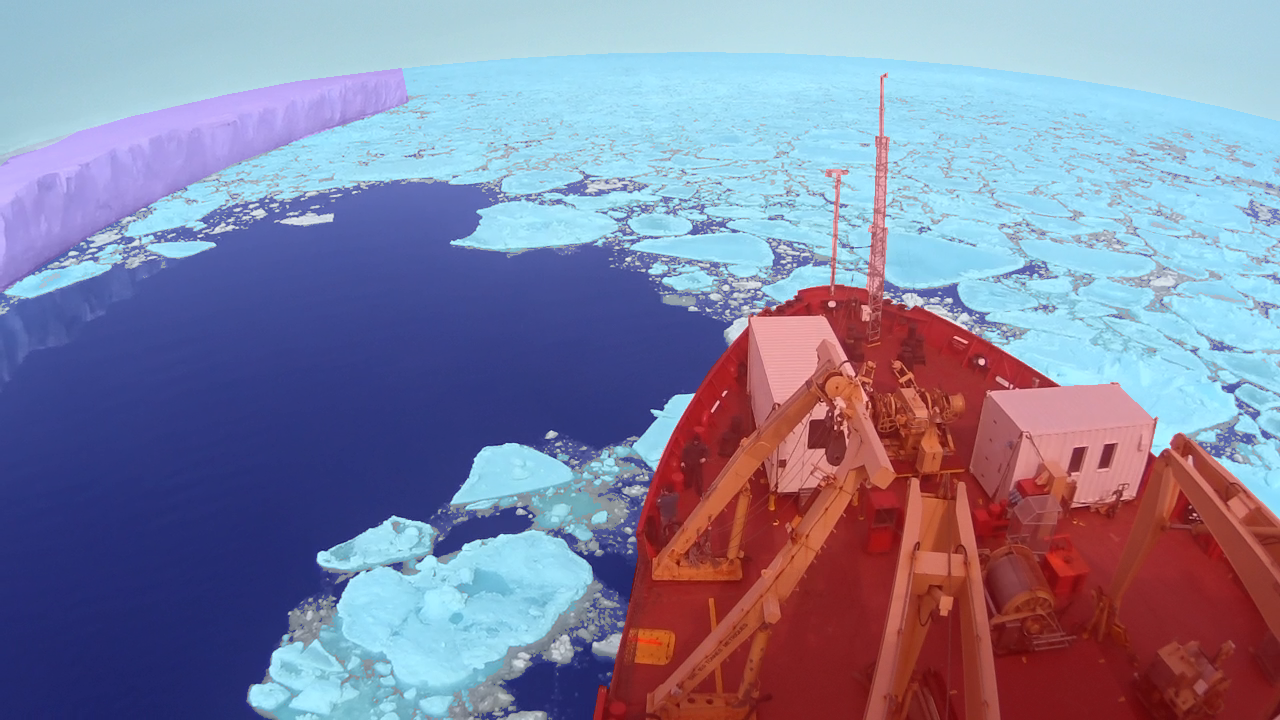}} &
    \ufc\rotcentered{\textbf{UPerFlow}} & 
    \ufc\begin{minipage}[c][3cm]{\linewidth}%
        \centering
        \suboverpic{test03uperflow}%
    \end{minipage} &%
    \ufc\begin{minipage}[c][3cm]{\linewidth}%
        \centering
        \suboverpic{occlusionlight03uperflow}%
    \end{minipage} &%
    \ufc\begin{minipage}[c][3cm]{\linewidth}%
        \centering
        \suboverpic{occlusionheavy03uperflow}%
    \end{minipage}\\%
    \rowcolor{blue!10} & \unc \rotcentered{UPerNet} &
    \unc\begin{minipage}[c][3cm]{\linewidth}%
        \centering
        \suboverpic{test43upernet}%
    \end{minipage} &%
    \unc\begin{minipage}[c][3cm]{\linewidth}%
        \suboverpic{occlusionlight43upernet}%
        \centering
    \end{minipage} &%
    \unc\begin{minipage}[c][3cm]{\linewidth}%
        \centering
        \suboverpic{occlusionheavy43upernet}%
    \end{minipage}\\%
    \rowcolor{blue!5}\multirow{-8}{*}{\includegraphics[width=.9\linewidth, valign=m]{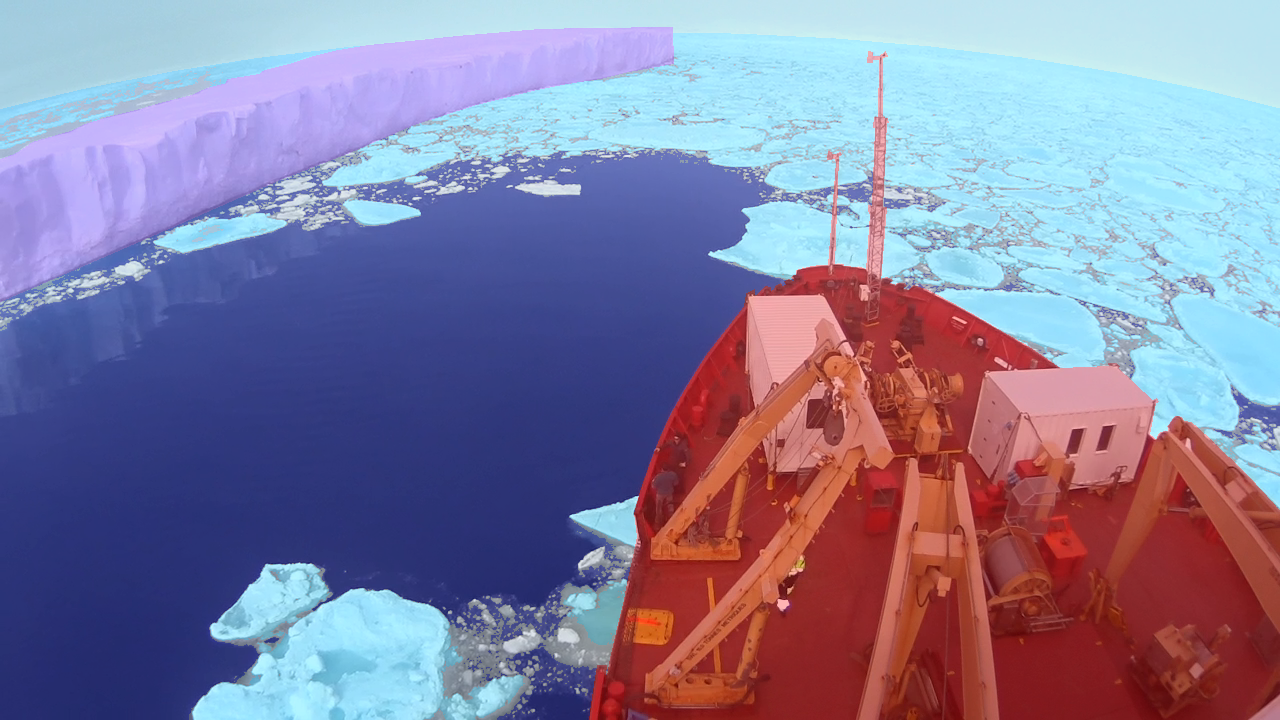}} &
    \ufc\rotcentered{\textbf{UPerFlow}} & 
    \ufc\begin{minipage}[c][3cm]{\linewidth}%
        \centering
        \suboverpic{test43uperflow}%
    \end{minipage} &%
    \ufc\begin{minipage}[c][3cm]{\linewidth}%
        \centering
        \suboverpic{occlusionlight43uperflow}%
    \end{minipage} &%
    \ufc\begin{minipage}[c][3cm]{\linewidth}%
        \centering
        \suboverpic{occlusionheavy43uperflow}%
    \end{minipage}\\%
    \rowcolor{blue!5} & \unc \rotcentered{UPerNet} &
    \unc\begin{minipage}[c][3cm]{\linewidth}%
        \centering
        \suboverpic{test83upernet}%
    \end{minipage} &%
    \unc\begin{minipage}[c][3cm]{\linewidth}%
        \centering
        \suboverpic{occlusionlight83upernet}%
    \end{minipage} &%
    \unc\begin{minipage}[c][3cm]{\linewidth}%
        \centering
        \suboverpic{occlusionheavy83upernet}%
    \end{minipage}\\%
    \rowcolor{blue!5}\multirow{-8}{*}{\includegraphics[width=.9\linewidth, valign=m]{test_overlay_003.png}} &
    \ufc\rotcentered{\textbf{UPerFlow}} & 
    \ufc\begin{minipage}[c][3cm]{\linewidth}%
        \centering
        \suboverpic{test83uperflow}%
    \end{minipage} &%
    \ufc\begin{minipage}[c][3cm]{\linewidth}%
        \centering
        \suboverpic{occlusionlight83uperflow}%
    \end{minipage} &%
    \ufc\begin{minipage}[c][3cm]{\linewidth}%
        \centering
        \suboverpic{occlusionheavy83uperflow}%
    \end{minipage}\\%
\end{tabular}
\caption{Example predictions for UPerNet (red) and UPerFlow (green) on the test dataset. The performance of the networks for different levels of occlusion are shown, next to the ground truth. UPerFlow performs accurately like UPerNet while reducing error around the iceberg and in occluded regions.}
\label{fig:test_results}
\end{sidewaysfigure*}

The performance of UPerNet and UPerFlow on both datasets is summarized in Table~\ref{tab:video_occlusion_results}.
Performance for the lightly occluded dataset is labeled ``Light'', while the heavily occluded dataset is labeled ``Heavy''. 
Since the occlusions impact only small portions of the images, only slight overall performance degradation is observed. 
However, assessing performance exclusively in the occluded regions highlights the networks’ actual capabilities under occlusion.
Despite training with similar artificial occlusions, UPerNet’s mIoU in these regions falls to 0.657 on the lightly occluded dataset and 0.535 on the heavily occluded one, marking a 30\% drop at the highest occlusion level. 
In contrast, UPerFlow’s video segmentation approach demonstrates stronger resilience: its mIoU decreases by less than 12\% on the lightly occluded dataset, achieving 0.727 mIoU, and slightly improves to 0.736 on the heavily occluded data. UPerFlow maintains stable performance regardless of occlusion severity, outperforming UPerNet by 37.6\% for worst-case occlusion.

\begin{figure*}[t]
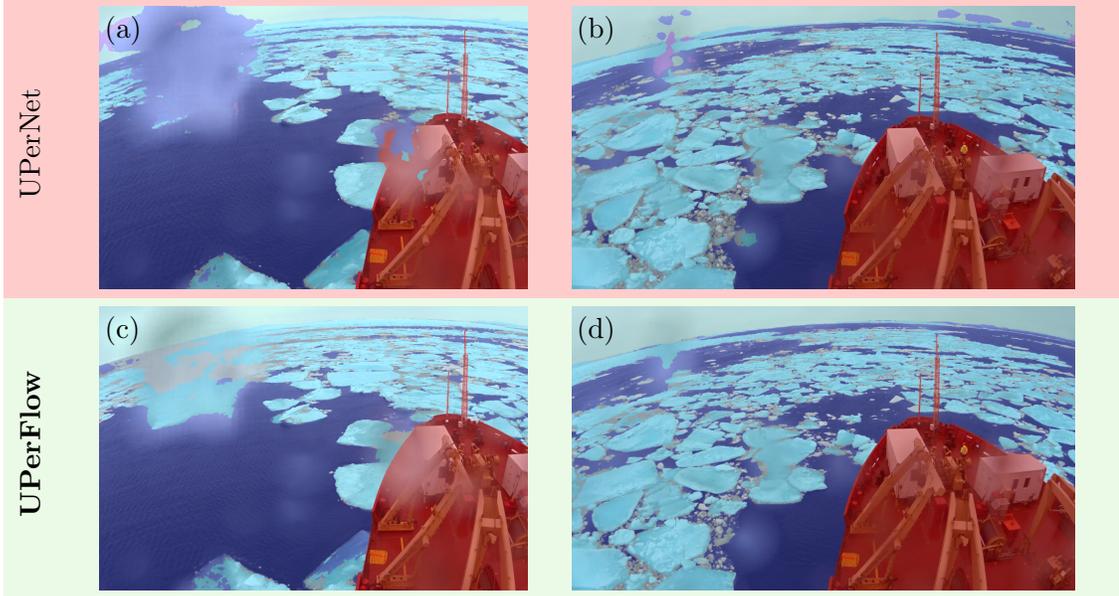

    \centering
    \begin{tabular}{c >{\centering\arraybackslash}p{0.45\linewidth} @{\hspace{-3em}}>{\centering\arraybackslash}p{0.45\linewidth}@{\hspace{-3em}}}
    \rowcolor{red!20}%
    \rotcentered{UPerNet} &%
    \begin{minipage}[c][4cm]{\linewidth}%
        \centering
        \suboverpic{occlusionreal104upernet}%
    \end{minipage} &%
    \begin{minipage}[c][4cm]{\linewidth}%
        \centering
        \suboverpic{occlusionreal196upernet}%
    \end{minipage}\\%
    \rowcolor{red!20!green!10}%
    \rotcentered{\textbf{UPerFlow}} &%
    \begin{minipage}[c][4cm]{\linewidth}%
        \centering
        \suboverpic{occlusionreal104uperflow}%
    \end{minipage} &%
    \begin{minipage}[c][4cm]{\linewidth}%
        \centering
        \suboverpic{occlusionreal196uperflow}%
    \end{minipage}\\%
    \end{tabular}
    \caption{Sample predictions of UPerNet (red) and UPerFlow (green) on unlabeled data with water droplets on the lens.}
    \label{fig:realocclusion}
\end{figure*}

Inference results of the networks on the occluded data are illustrated in Figure~\ref{fig:test_results}. 
UPerNet performs mostly well on the lightly occluded data, effectively utilizing partially visible features to generate reasonable predictions in occluded regions. 
However, it struggles with iceberg predictions, producing spotty and inconsistent results, as observed in Figure~\ref{fig:occlusionlight43upernet}. 
This is likely due to the infrequent appearance of iceberg in the training data, especially from the new perspective presented in the test data. 
These findings indicate that even when partial information is available through occlusions, the predictive capability of single-image networks is constrained by their learned ability to interpolate shapes.

The trend of spotty and inconsistent results is exacerbated on the heavily occluded data.
Without any discernible features in the occluded regions, the network is forced to guess the shape and location of objects, leading to significant inaccuracies. 
For instance, in Figures~\ref{fig:occlusionheavy03upernet}~and~\ref{fig:occlusionheavy83upernet}, the predicted shape of the ship in the occluded region is severely distorted from what would be expected.our
The network also produces large, coarse ice floe predictions, as it lacks specific knowledge of individual ice floe shapes.
Additionally, in Figures~\ref{fig:occlusionheavy03upernet}~and~\ref{fig:occlusionheavy43upernet}, the network tends to predict icebergs in occluded regions, even when not near an iceberg, demonstrating the unreliability of the network's results in occluded regions.

In contrast, the predictions from UPerFlow remain consistent, regardless of level of occlusion.
The network generates plausible predictions for object boundaries, and avoids creating random iceberg predictions, ensuring that all predictions are reasonable approximations of the occluded areas.
These results underscore the advantage of video segmentation networks over image segmentation models, as they can utilize temporal information to more effectively infer details in occluded regions.

Finally, Figure~\ref{fig:realocclusion} illustrates the networks' performance when faced with real occlusion, such as water droplets on the lens.
UPerNet predictions are highlighted in red and those of UPerFlow are highlighted in green.
Despite being trained with random erasing, UPerNet struggles with these real-world occlusions, producing nearly random predictions in the affected areas.
Similar to the iceberg scenario, the data behind the occluded regions is unfamiliar to the model, leaving the network unable to accurately interpolate the shape of the obscured objects.
In contrast, UPerFlow creates reasonable predictions in occluded regions caused by water droplets, demonstrating its superior ability to handle real occlusions compared to traditional image segmentation approaches.

Accurate predictions in occluded regions are essential for safe navigation and decision-making in ice-covered waters. 
When small or partial occlusions obscure parts of the visual field, they may hide critical objects, such as floating icebergs or hazardous ice floes, that pose severe risks. 
For instance, an iceberg hidden behind a minor obstruction could result in the misjudgment of distance or impact force, significantly compromising safety. Therefore, robust occlusion handling ensures that these potential hazards are detected, even when visibility is limited, enhancing the reliability of automated detection systems in high-stakes environments.

\section{Conclusion} \label{conclusion}
The need for more accurate and efficient methods to segment close-range sea ice imagery motivated our key contributions that include a novel labeling strategy and a modified video segmentation architectures. 
The labeling strategy allows fine annotations of high-resolution optical imagery into ice floes, brash ice, water, ships, sky, and icebergs, and provides a foundation for future studies. 
The video segmentation strategy demonstrated that by embedding optical flow features directly into the network, performance can be improved overall and significantly in occluded regions.
The approach achieved an overall 94.8\% accuracy and mIoU of 0.844.
On artificially occluded data, the method performs consistently regardless of level of occlusion, and outperforms the image segmentation model by roughly 40\%, with an mIoU of 0.736 in heavily occluded regions. Additionally, the approach suffers significantly less from misclassifications when tested on imagery with real data artifacts, such as reflections in the water and water droplets on the lens. Before proposing our modified video segmentation approach, we also assessed existing image semantic segmentation networks and determined that UPerNet outperforms the other architectures achieving 94.6\% accuracy and an mIoU of 0.838. Despite its strong performance, UPerNet still encountered challenges with artifacts common in realistic sea ice data, which the proposed video segmentation model was able to overcome.

The contributions of this study collectively highlight the potential of neural networks to provide robust ice condition assessments and support researchers and navigators in making objective, data-driven decisions. By enhancing the safety and efficiency of navigation in the increasingly accessible Arctic waters, this research supports the broader goal of improving decision-making capabilities in this critical and evolving region.

\section*{Acknowledgments}
This work is partially supported by the National Research Council (Ocean Supercluster Support Program) and 
National Research Council/University of Waterloo Collaboration Centre (NUCC).
We would like to extend our gratitude to Matthew Garvin and Robert Gash from the National Research Council Canada.
Their collaborative efforts in collecting and sharing the Amundsen data, along with their expertise and ongoing support were invaluable to the development of the present work.  
We would also like to thank Richard Duan and JD Zhu for their respective contributions to this project.

\bibliographystyle{ieeetr} 
\bibliography{references}

\begin{thebibliography}{10}

\bibitem{NVIDIAkeynote}
``{GTC November 2021 Keynote with NVIDIA CEO Jensen Huang 2021},'' 11 2021.

\bibitem{2060iceconditions}
J.~C. Stroeve, V.~Kattsov, A.~Barrett, M.~Serreze, T.~Pavlova, M.~Holland, and
  W.~N. Meier, ``{Trends in Arctic sea ice extent from CMIP5, CMIP3 and
  observations},'' {\em Geophysical Research Letters}, vol.~39, no.~16, 2012.

\bibitem{ostreng2013shipping}
W.~Ostreng, K.~M. Eger, B.~Fl{\o}istad, A.~J{\o}rgensen-Dahl, L.~Lothe,
  M.~Mejl{\ae}nder-Larsen, and T.~Wergeland, {\em {Shipping in Arctic waters: a
  comparison of the Northeast, Northwest and trans polar passages}}.
\newblock Springer Science \& Business Media, 2013.

\bibitem{garvin2020review}
M.~Garvin, ``{Review of technologies for real-time shipboard ice severity
  sensing},'' in {\em International Conference on Smart Ship Technology}, 2020.

\bibitem{lund2018radar}
B.~Lund, H.~C. Graber, P.~O.~G. Persson, M.~Smith, M.~Doble, J.~Thomson, and
  P.~Wadhams, ``{Arctic Sea Ice Drift Measured by Shipboard Marine Radar},''
  {\em Journal of Geophysical Research: Oceans}, vol.~123, no.~6,
  pp.~4298--4321, 2018.

\bibitem{heyn2020accel}
H.-M. Heyn, M.~Blanke, and R.~Skjetne, ``{Ice Condition Assessment Using
  Onboard Accelerometers and Statistical Change Detection},'' {\em IEEE Journal
  of Oceanic Engineering}, vol.~45, no.~3, pp.~898--914, 2020.

\bibitem{petty2016lidar}
A.~A. Petty, M.~C. Tsamados, N.~T. Kurtz, S.~L. Farrell, T.~Newman, J.~P.
  Harbeck, D.~L. Feltham, and J.~A. Richter-Menge, ``{Characterizing Arctic sea
  ice topography using high-resolution IceBridge data},'' {\em The Cryosphere},
  vol.~10, no.~3, pp.~1161--1179, 2016.

\bibitem{remund1998kmeans}
Q.~Remund, D.~Long, and M.~Drinkwater, ``{Polar sea-ice classification using
  enhanced resolution NSCAT data},'' in {\em IGARSS'98. Sensing and Managing
  the Environment. 1998 IEEE International Geoscience and Remote Sensing.
  Symposium Proceedings.(Cat. No. 98CH36174)}, vol.~4, pp.~1976--1978, IEEE,
  1998.

\bibitem{coggins2014kmeans}
J.~H. Coggins, A.~J. McDonald, and B.~Jolly, ``{Synoptic climatology of the
  Ross Ice Shelf and Ross Sea region of Antarctica: k-means clustering and
  validation.},'' {\em International journal of climatology}, vol.~34, no.~7,
  2014.

\bibitem{fuvckar2016kmeans}
N.~S. Fu{\v{c}}kar, V.~Guemas, N.~C. Johnson, F.~Massonnet, and F.~J.
  Doblas-Reyes, ``{Clusters of interannual sea ice variability in the northern
  hemisphere},'' {\em Climate Dynamics}, vol.~47, pp.~1527--1543, 2016.

\bibitem{sobiech2013kmeans}
J.~Sobiech and W.~Dierking, ``{Observing lake-and river-ice decay with SAR:
  advantages and limitations of the unsupervised k-means classification
  approach},'' {\em Annals of Glaciology}, vol.~54, no.~62, pp.~65--72, 2013.

\bibitem{bharathi2013kmeans}
P.~Bharathi and P.~Subashini, ``{Texture based color segmentation for infrared
  river ice images using K-means clustering},'' in {\em 2013 International
  Conference on Signal Processing, Image Processing \& Pattern Recognition},
  pp.~298--302, IEEE, 2013.

\bibitem{WEISSLING2009eiscam}
B.~Weissling, S.~Ackley, P.~Wagner, and H.~Xie, ``{EISCAM — Digital image
  acquisition and processing for sea ice parameters from ships},'' {\em Cold
  Regions Science and Technology}, vol.~57, no.~1, pp.~49--60, 2009.

\bibitem{SANDRU2020kmeans}
A.~Sandru, H.~Hyyti, A.~Visala, and P.~Kujala, ``{A Complete Process For
  Shipborne Sea-Ice Field Analysis Using Machine Vision},'' {\em
  IFAC-PapersOnLine}, vol.~53, no.~2, pp.~14539--14545, 2020.
\newblock 21st IFAC World Congress.

\bibitem{zhang2014otsukmeans}
Q.~Zhang and R.~Skjetne, ``{Image techniques for identifying sea-ice
  parameters},'' in {\em Modeling, Identification and Control}, vol.~35,
  pp.~293--301, 2014.

\bibitem{desilva2016iceunsupervised}
O.~De~Silva, C.~Daley, J.~R. Dolny, G.~K. Mann, R.~G. Gosine, and D.~Peters,
  ``{Vision analysis of pack ice for potential use in a hazard warning and
  avoidance system},'' tech. rep., Memorial University of Newfoundland, 2016.

\bibitem{wenjun2016unsupervisedconcentration}
W.~Lu, Q.~Zhang, R.~Lubbad, S.~Løset, and R.~Skjetne, ``{A Shipborne
  Measurement System to Acquire Sea Ice Thickness and Concentration at
  Engineering Scale},'' vol.~All Days of {\em OTC Arctic Technology
  Conference}, pp.~OTC--27361--MS, 10 2016.

\bibitem{kalke2018svm}
H.~Kalke and M.~Loewen, ``{Support vector machine learning applied to digital
  images of river ice conditions},'' {\em Cold Regions Science and Technology},
  vol.~155, pp.~225--236, 2018.

\bibitem{li2020cnnsurvey}
Z.~Li, F.~Liu, W.~Yang, S.~Peng, and J.~Zhou, ``{A Survey of Convolutional
  Neural Networks: Analysis, Applications, and Prospects},'' {\em IEEE
  Transactions on Neural Networks and Learning Systems}, vol.~33, no.~12,
  pp.~6999--7019, 2022.

\bibitem{dowden}
B.~Dowden, O.~De~Silva, and W.~Huang, ``{Sea Ice Image Semantic Segmentation
  Using Deep Neural Networks},'' in {\em Global Oceans 2020: Singapore – U.S.
  Gulf Coast}, pp.~1--5, 2020.

\bibitem{icerainremoval}
N.~M. Alsharay, Y.~Chen, O.~A. Dobre, and O.~De~Silva, ``{Improved Sea-Ice
  Identification Using Semantic Segmentation With Raindrop Removal},'' {\em
  IEEE Access}, vol.~10, pp.~21599--21607, 2022.

\bibitem{icegan}
N.~M. Alsharay, O.~A. Dobre, Y.~Chen, and O.~De~Silva, ``{Sea-Ice
  Classification Using Conditional Generative Adversarial Networks},'' {\em
  IEEE Sensors Letters}, vol.~7, no.~4, pp.~1--4, 2023.

\bibitem{dowdenfaster}
N.~Balasooriya, B.~Dowden, J.~Chen, O.~De~Silva, and W.~Huang, ``{In-situ Sea
  Ice Detection using DeepLabv3 Semantic Segmentation},'' in {\em OCEANS 2021:
  San Diego – Porto}, pp.~1--7, 2021.

\bibitem{icedeeplabattention}
S.~Li, M.~Wang, J.~Wu, S.~Sun, M.~Shi, and R.~Ma, ``{Sea ice detection network
  for icebreakers in polar environments with attention-based deeplabv3+
  architecture},'' {\em Journal of Physics: Conference Series}, vol.~2718,
  p.~012062, 3 2024.

\bibitem{icedeeplabattention2}
C.~Zhang, X.~Chen, and S.~Ji, ``{Semantic image segmentation for sea ice
  parameters recognition using deep convolutional neural networks},'' {\em
  International Journal of Applied Earth Observation and Geoinformation},
  vol.~112, p.~102885, 2022.

\bibitem{iceunetmodifications}
J.~Zhao, L.~Chen, J.~Li, and Y.~Zhao, ``{Semantic Segmentation of Sea Ice Based
  on U-net Network Modification},'' in {\em 2022 IEEE International Conference
  on Robotics and Biomimetics (ROBIO)}, pp.~1151--1156, 2022.

\bibitem{panchi}
N.~Panchi, E.~Kim, and A.~Bhattacharyya, ``{Supplementing Remote Sensing of
  Ice: Deep Learning-Based Image Segmentation System for Automatic Detection
  and Localization of Sea-ice Formations From Close-Range Optical Images},''
  {\em IEEE Sensors Journal}, vol.~21, no.~16, pp.~18004--18019, 2021.

\bibitem{panchi2024weatherdegraded}
N.~Panchi and E.~Kim, ``{Deep Learning Strategies for Analysis of
  Weather-Degraded Optical Sea Ice Images},'' {\em IEEE Sensors Journal},
  vol.~24, no.~9, pp.~15252--15272, 2024.

\bibitem{nathanielbpalmer}
C.~Brooks, ``{Two Months Breaking Ice (In Under Five Minutes)},'' 5 2013.

\bibitem{otsu}
N.~Otsu, ``{A Threshold Selection Method from Gray-Level Histograms},'' {\em
  IEEE Transactions on Systems, Man, and Cybernetics}, vol.~9, no.~1,
  pp.~62--66, 1979.

\bibitem{arora2008multilevel}
S.~Arora, J.~Acharya, A.~Verma, and P.~K. Panigrahi, ``{Multilevel thresholding
  for image segmentation through a fast statistical recursive algorithm},''
  {\em Pattern Recognition Letters}, vol.~29, no.~2, pp.~119--125, 2008.

\bibitem{haverkamp1995localdynamicthresholding}
D.~Haverkamp, L.~K. Soh, and C.~Tsatsoulis, ``{A comprehensive, automated
  approach to determining sea ice thickness from SAR data},'' {\em IEEE
  Transactions on Geoscience and Remote Sensing}, vol.~33, no.~1, pp.~46--57,
  1995.

\bibitem{lu2010shipthresholding}
P.~Lu and Z.~Li, ``{A Method of Obtaining Ice Concentration and Floe Size From
  Shipboard Oblique Sea Ice Images},'' {\em IEEE Transactions on Geoscience and
  Remote Sensing}, vol.~48, no.~7, pp.~2771--2780, 2010.

\bibitem{long2015fcn}
J.~Long, E.~Shelhamer, and T.~Darrell, ``{Fully Convolutional Networks for
  Semantic Segmentation},'' in {\em Proceedings of the IEEE Conference on
  Computer Vision and Pattern Recognition (CVPR)}, 6 2015.

\bibitem{ronneberger2015unet}
O.~Ronneberger, P.~Fischer, and T.~Brox, ``U-net: Convolutional networks for
  biomedical image segmentation,'' in {\em Medical Image Computing and
  Computer-Assisted Intervention -- MICCAI 2015} (N.~Navab, J.~Hornegger, W.~M.
  Wells, and A.~F. Frangi, eds.), (Cham), pp.~234--241, Springer International
  Publishing, 2015.

\bibitem{chen2018deeplabv3plus}
L.-C. Chen, Y.~Zhu, G.~Papandreou, F.~Schroff, and H.~Adam, ``Encoder-decoder
  with atrous separable convolution for semantic image segmentation,'' in {\em
  Computer Vision -- ECCV 2018} (V.~Ferrari, M.~Hebert, C.~Sminchisescu, and
  Y.~Weiss, eds.), (Cham), pp.~833--851, Springer International Publishing,
  2018.

\bibitem{zhao2017pspnet}
H.~Zhao, J.~Shi, X.~Qi, X.~Wang, and J.~Jia, ``{ Pyramid Scene Parsing Network
  },'' in {\em 2017 IEEE Conference on Computer Vision and Pattern Recognition
  (CVPR)}, (Los Alamitos, CA, USA), pp.~6230--6239, IEEE Computer Society, July
  2017.

\bibitem{xiao2018upernet}
T.~Xiao, Y.~Liu, B.~Zhou, Y.~Jiang, and J.~Sun, ``{Unified Perceptual Parsing
  for Scene Understanding},'' in {\em Proceedings of the European Conference on
  Computer Vision (ECCV)}, 9 2018.

\bibitem{FarnebackOpticalFlow}
G.~Farneb{\"a}ck, ``Two-frame motion estimation based on polynomial
  expansion,'' in {\em Image Analysis} (J.~Bigun and T.~Gustavsson, eds.),
  (Berlin, Heidelberg), pp.~363--370, Springer Berlin Heidelberg, 2003.

\bibitem{lucas1981opticalflow}
B.~D. Lucas and T.~Kanade, ``{An iterative image registration technique with an
  application to stereo vision},'' in {\em IJCAI'81: 7th international joint
  conference on Artificial intelligence}, vol.~2, pp.~674--679, 1981.

\bibitem{horn1981opticalflow}
B.~K. Horn and B.~G. Schunck, ``{Determining optical flow},'' {\em Artificial
  intelligence}, vol.~17, no.~1-3, pp.~185--203, 1981.

\bibitem{ilg2016flownet2}
E.~Ilg, N.~Mayer, T.~Saikia, M.~Keuper, A.~Dosovitskiy, and T.~Brox, ``{FlowNet
  2.0: Evolution of Optical Flow Estimation with Deep Networks},'' in {\em 2017
  IEEE Conference on Computer Vision and Pattern Recognition (CVPR)},
  pp.~1647--1655, 2017.

\bibitem{sun2018pwcnet}
D.~Sun, X.~Yang, M.-Y. Liu, and J.~Kautz, ``{PWC-Net: CNNs for Optical Flow
  Using Pyramid, Warping, and Cost Volume},'' in {\em 2018 IEEE/CVF Conference
  on Computer Vision and Pattern Recognition}, pp.~8934--8943, 2018.

\bibitem{yang2019hsmnet}
G.~Yang, J.~Manela, M.~Happold, and D.~Ramanan, ``{Hierarchical Deep Stereo
  Matching on High-Resolution Images},'' in {\em 2019 IEEE/CVF Conference on
  Computer Vision and Pattern Recognition (CVPR)}, pp.~5510--5519, 2019.

\bibitem{teed2020raft}
Z.~Teed and J.~Deng, ``{RAFT: Recurrent all-pairs field transforms for optical
  flow},'' in {\em Computer Vision--ECCV 2020: 16th European Conference,
  Glasgow, UK, August 23--28, 2020, Proceedings, Part II 16}, pp.~402--419,
  Springer, 2020.

\bibitem{nilsson2017gru}
D.~Nilsson and C.~Sminchisescu, ``{Semantic video segmentation by gated
  recurrent flow propagation},'' in {\em Proceedings of the IEEE conference on
  computer vision and pattern recognition}, pp.~6819--6828, 2018.

\bibitem{3dcrf}
A.~Kundu, V.~Vineet, and V.~Koltun, ``{Feature Space Optimization for Semantic
  Video Segmentation},'' in {\em 2016 IEEE Conference on Computer Vision and
  Pattern Recognition (CVPR)}, pp.~3168--3175, 2016.

\bibitem{chandra2018crf}
S.~Chandra, C.~Couprie, and I.~Kokkinos, ``{Deep spatio-temporal random fields
  for efficient video segmentation},'' in {\em Proceedings of the IEEE
  conference on Computer Vision and Pattern Recognition}, pp.~8915--8924, 2018.

\bibitem{gadde2017netwarp}
R.~Gadde, V.~Jampani, and P.~V. Gehler, ``{Semantic video CNNs through
  representation warping},'' in {\em Proceedings of the IEEE International
  Conference on Computer Vision}, pp.~4453--4462, 2017.

\bibitem{ding2019efc}
M.~Ding, Z.~Wang, B.~Zhou, J.~Shi, Z.~Lu, and P.~Luo, ``{Every frame counts:
  Joint learning of video segmentation and optical flow},'' in {\em Proceedings
  of the AAAI conference on artificial intelligence}, vol.~34,
  pp.~10713--10720, 2020.

\bibitem{liu2020etc}
Y.~Liu, C.~Shen, C.~Yu, and J.~Wang, ``{Efficient Semantic Video Segmentation
  with Per-frame Inference},'' {\em ECCV}, 2020.

\bibitem{huang2018videouncertainty}
P.-Y. Huang, W.-T. Hsu, C.-Y. Chiu, T.-F. Wu, and M.~Sun, ``{Efficient
  uncertainty estimation for semantic segmentation in videos},'' in {\em
  Proceedings of the European Conference on Computer Vision (ECCV)},
  pp.~520--535, 2018.

\bibitem{Chengsegflow}
J.~Cheng, Y.-H. Tsai, S.~Wang, and M.-H. Yang, ``{SegFlow: Joint Learning for
  Video Object Segmentation and Optical Flow},'' in {\em IEEE International
  Conference on Computer Vision (ICCV)}, 2017.

\bibitem{resnet}
K.~He, X.~Zhang, S.~Ren, and J.~Sun, ``{Deep Residual Learning for Image
  Recognition},'' 2015.

\bibitem{fischer2015flownet}
A.~Dosovitskiy, P.~Fischer, E.~Ilg, P.~Hausser, C.~Hazirbas, V.~Golkov, P.~Van
  Der~Smagt, D.~Cremers, and T.~Brox, ``{Flownet: Learning optical flow with
  convolutional networks},'' in {\em Proceedings of the IEEE international
  conference on computer vision}, pp.~2758--2766, 2015.

\bibitem{zhu2019vplr}
Y.~Zhu, K.~Sapra, F.~A. Reda, K.~J. Shih, S.~Newsam, A.~Tao, and B.~Catanzaro,
  ``{Improving semantic segmentation via video propagation and label
  relaxation},'' in {\em Proceedings of the IEEE/CVF conference on computer
  vision and pattern recognition}, pp.~8856--8865, 2019.

\bibitem{WMO}
``{Sea-Ice Nomenclature},'' tech. rep., World Meteorological Organization
  (WMO), 2014.

\bibitem{cameraCalibration}
F.~Remondino and C.~Fraser, ``{Digital camera calibration methods:
  considerations and comparisons},'' {\em International Archives of the
  Photogrammetry, Remote Sensing and Spatial Information Sciences}, vol.~36,
  no.~5, pp.~266--272, 2006.

\bibitem{coordinateSystem}
P.~Moon and D.~E. Spencer, {\em {Field theory handbook: including coordinate
  systems, differential equations and their solutions}}.
\newblock Springer, 2012.

\bibitem{cvat}
B.~Sekachev, N.~Manovich, M.~Zhiltsov, A.~Zhavoronkov, D.~Kalinin, B.~Hoff,
  TOsmanov, D.~Kruchinin, A.~Zankevich, DmitriySidnev, M.~Markelov,
  Johannes222, M.~Chenuet, a~andre, telenachos, A.~Melnikov, J.~Kim, L.~Ilouz,
  N.~Glazov, Priya4607, R.~Tehrani, S.~Jeong, V.~Skubriev, S.~Yonekura, vugia
  truong, zliang7, lizhming, and T.~Truong, ``{CVAT},'' Aug. 2020.

\bibitem{optuna_2019}
T.~Akiba, S.~Sano, T.~Yanase, T.~Ohta, and M.~Koyama, ``{Optuna: A
  Next-generation Hyperparameter Optimization Framework},'' in {\em Proceedings
  of the 25th {ACM} {SIGKDD} International Conference on Knowledge Discovery
  and Data Mining}, 2019.

\bibitem{zhong2017randomerasing}
Z.~Zhong, L.~Zheng, G.~Kang, S.~Li, and Y.~Yang, ``{Random erasing data
  augmentation},'' in {\em Proceedings of the AAAI conference on artificial
  intelligence}, vol.~34, pp.~13001--13008, 2020.

\bibitem{chen2017deeplabv3}
L.-C. Chen, G.~Papandreou, F.~Schroff, and H.~Adam, ``{Rethinking Atrous
  Convolution for Semantic Image Segmentation},'' 2017.

\bibitem{loshchilov2019adamw}
I.~Loshchilov, F.~Hutter, {\em et~al.}, ``{Fixing weight decay regularization
  in adam},'' {\em arXiv preprint arXiv:1711.05101}, vol.~5, 2017.

\bibitem{loshchilov2017cosinelr}
I.~Loshchilov and F.~Hutter, ``{{SGDR}: Stochastic Gradient Descent with Warm
  Restarts},'' in {\em International Conference on Learning Representations},
  2017.

\bibitem{shorten2019dataaugmentation}
C.~Shorten and T.~M. Khoshgoftaar, ``{A survey on image data augmentation for
  deep learning},'' {\em Journal of big data}, vol.~6, no.~1, pp.~1--48, 2019.

\bibitem{szegedy2014googlenet}
C.~Szegedy, W.~Liu, Y.~Jia, P.~Sermanet, S.~Reed, D.~Anguelov, D.~Erhan,
  V.~Vanhoucke, and A.~Rabinovich, ``{Going deeper with convolutions},'' in
  {\em Proceedings of the IEEE conference on computer vision and pattern
  recognition}, pp.~1--9, 2015.

\bibitem{howard2013improvements}
A.~G. Howard, ``{Some improvements on deep convolutional neural network based
  image classification},'' {\em arXiv preprint arXiv:1312.5402}, 2013.

\end{thebibliography}

\cleardoublepage
\section{APPENDICES}
\appendix
\setcounter{figure}{0}
\setcounter{table}{0}
\renewcommand\thefigure{\Alph{section}.\arabic{figure}}
\renewcommand\thetable{\Alph{section}.\arabic{table}}
\section{Labeling Implementation} \label{labelingdetails}
\subsection{Ice Floe Annotation}
For the video in Figure~\ref{fig:labelingProcess}, three regions were defined using extrinsic parameter value of $Y^*=3$ and intrinsic parameter value of $a=800$. 
In each region, the top 30th percentile of brightest classes chosen by \texttt{multithresh} are chosen as ice floes. 
Ice floe clusters are then identified by \texttt{bwareafilt}, where clusters less than 700 pixels in area are removed from the ice floe classification.
The range of thresholds to label water was chosen as less than 100 in the near-field and 340 in the far-field. 
These values were found to be fairly consistent across sequential frames, however the number may vary by video. 
The labeling hyperparameters are summarized in Table~\ref{tab:labelingvalues}.

\begin{table}[h!]
    \centering
    \begin{tabular}{|c|c|} \hline
         \textbf{Hyperparameter} & \textbf{Value} \\ \hline
         Number of Regions & 3 \\ \hline
         $Y^*$ & 3 \\ \hline
         a & 800 \\ \hline
         Ice floe brightest percentile & 30th \\ \hline
         Water threshold range & [100, 340] \\ \hline
        \end{tabular}
    \caption{The labeling hyperparameters used for semi-manual annotation.}
    \label{tab:labelingvalues}
\end{table}

\section{Model Implementation} \label{modeldetails}
\setcounter{figure}{0}
\setcounter{table}{0}
\subsection{Training Hyperparameters}
A large Bayesian hyperparameter search using Optuna~\cite{optuna_2019} was conducted to simultaneously optimize the selection of image segmentation network, along with training and data augmentation hyperparameters. 
The tested networks included a simple fully convolutional network~\cite{long2015fcn}, DeepLabV3~\cite{chen2017deeplabv3}, DeepLabV3+~\cite{chen2018deeplabv3plus}, UNet~\cite{ronneberger2015unet}, UPerNet~\cite{xiao2018upernet}, and PSPNet~\cite{zhao2017pspnet}.
The best-performing model was identified as UPerNet, with optimal training and data augmentation hyperparameters listed in Tables~\ref{tab:traininghpvalues}~and~\ref{tab:dataaugmentationvalues}, respectively. 
All models were trained for up to 800 epochs, using an AdamW optimizer~\cite{loshchilov2019adamw} with a cosine annealing learning rate scheduler~\cite{loshchilov2017cosinelr}. Early stopping was applied if validation loss failed to improve over 150 epochs, and the best-performing model checkpoint was saved for evaluation. The optimal learning rate, between 1E-4 and 1E-6, and a weight decay of 1E-6 were identified through Optuna.

\begin{table}[h!]
    \centering
    \begin{tabular}{|>{\raggedright\arraybackslash}m{0.5\linewidth}|>{\centering\arraybackslash}m{0.5\linewidth}|} \hline
     \textbf{Hyperparameter} & \textbf{Value} \\ \hline
     Learning rate optimizer & AdamW~\cite{loshchilov2019adamw} \\ \hline
     Learning rate scheduler & Cosine annealing with warm restarts~\cite{loshchilov2017cosinelr} \\
     - Number of iterations until restart & 10 \\
     - Restart period factor & 2 \\ 
     - Learning rate range & [1E-4, 1E-6] \\ \hline
     Weight decay & 1E-6 \\ \hline
     Maximum number of epochs & 800 \\ \hline
     Number of overfit epochs & 150 \\ \hline
     \end{tabular}
    \caption{Hyperparameters used during training.}
    \label{tab:traininghpvalues}
\end{table}

\subsection{Data Augmentation Hyperparameters}
The data augmentations~\cite{shorten2019dataaugmentation} used for model training are summarized in Table~\ref{tab:dataaugmentationvalues}.
These data augmentations were selected to increase the variance of the dataset while remaining true to the variations expected in the Arctic environment. The hyperparameter values for random erasing were not included in the hyperparameter search, instead opting to use those suggested by Zhong et al.~\cite{zhong2017randomerasing}.
\begin{table}[h!]
    \centering
    \begin{tabular}{|l|c|} \hline
     \textbf{Hyperparameter} & \textbf{Value} \\ \hline
     Random color jittering~\cite{szegedy2014googlenet, howard2013improvements} & (-) \\
     - Random saturation jittering range & [0.2, 1.8] \\
     - Random contrast jittering range & [0.5, 1.5] \\ \hline
     Random rotation range & $[-5\degree, 5\degree]$ \\ \hline
     Random Gaussian blur & (-) \\
     - Kernel size & 5 \\
     - Standard deviation range & $[0.5, 2]$ \\ \hline
     Random horizontal flipping probability & 0.5 \\ \hline
     Random crop size & $[512, 512]$ \\ \hline
     Mean and Gaussian random erasing~\cite{zhong2017randomerasing} & (-) \\ 
     - Area ratio range & $(0.02, 0.33)$ \\
     - Aspect ratio range & $(0.3, 2.5)$ \\ \hline
     \end{tabular}
    \caption{Data augmentation hyperparameters used during training before normalization to the mean and standard deviation of the ImageNet dataset.}
    \label{tab:dataaugmentationvalues}
\end{table}

\end{document}